\documentclass[10pt,journal,compsoc]{IEEEtran}

\ifCLASSOPTIONcompsoc
  \usepackage[nocompress]{cite}
  \usepackage{graphicx}
\else
  \usepackage{cite}
\fi
\ifCLASSINFOpdf
\else
\fi

\usepackage{subfigure}
\usepackage{multirow}

\begin{document}
%
\title{T-GCN: A Temporal Graph Convolutional Network for Traffic Prediction}
%
%
%
%

\author{Ling Zhao, Yujiao Song, Chao Zhang, Yu Liu, Pu Wang, Tao Lin, Min Deng and~Haifeng~Li,~\IEEEmembership{Member,~IEEE}
\IEEEcompsocitemizethanks{\IEEEcompsocthanksitem L. Zhao, Y. Song, M. Deng, and H. Li are with School of Geosciences and Info-Physics, Central South University, Changsha 410083, China.\protect\\
Corresponding author: H.Li lihaifeng@csu.edu.cn
\IEEEcompsocthanksitem C. Zhang are with School of Computational Science and Engineering, Georgia Institute of Technology, Atlanta, GA 30332, USA.
\IEEEcompsocthanksitem Y. Liu are with with Institute of Remote Sensing and Geographic Information System, Peking University, Beijing, China.
\IEEEcompsocthanksitem W. Pu are with with School of Traffic Transportation Engineering, Central South University, Changsha 410083, China.
\IEEEcompsocthanksitem L. Tao are with College of Biosystems Engineering and Food Science, Zhejiang University, Hangzhou, China. \protect \\ Preprint. Work in progress.}
}

%
%

\markboth{Journal of \LaTeX\ Class Files,~Vol.~14, No.~8, August~2015}%
{Shell \MakeLowercase{\textit{et al.}}: Bare Advanced Demo of IEEEtran.cls for IEEE Computer Society Journals}
%



\IEEEtitleabstractindextext{%
\begin{abstract}
Accurate and real-time traffic forecasting plays an important role in the Intelligent Traffic System and is of great significance for urban traffic planning, traffic management, and traffic control. However, traffic forecasting has always been considered an “open” scientific issue, owing to the constraints of urban road network topological structure and the law of dynamic change with time, namely, spatial dependence and temporal dependence. To capture the spatial and temporal dependence simultaneously, we propose a novel neural network-based traffic forecasting method, the temporal graph convolutional network (T-GCN) model, which is in combination with the graph convolutional network (GCN) and gated recurrent unit (GRU). Specifically, the GCN is used to learn complex topological structures to capture spatial dependence and the gated recurrent unit is used to learn dynamic changes of traffic data to capture temporal dependence. Then, the T-GCN model is employed to traffic forecasting based on the urban road network. Experiments demonstrate that our T-GCN model can obtain the spatio-temporal correlation from traffic data and the predictions outperform state-of-art baselines on real-world traffic datasets. Our tensorflow implementation of the T-GCN is available at https://github.com/lehaifeng/T-GCN.
\end{abstract}

}

\maketitle

\IEEEdisplaynontitleabstractindextext

%
\IEEEpeerreviewmaketitle

\ifCLASSOPTIONcompsoc
\IEEEraisesectionheading{\section{Introduction}\label{sec:introduction}}
\else
\section{Introduction}
\label{sec:introduction}
\fi

\begin{figure}
	\centering
	\includegraphics[width=0.9\linewidth]{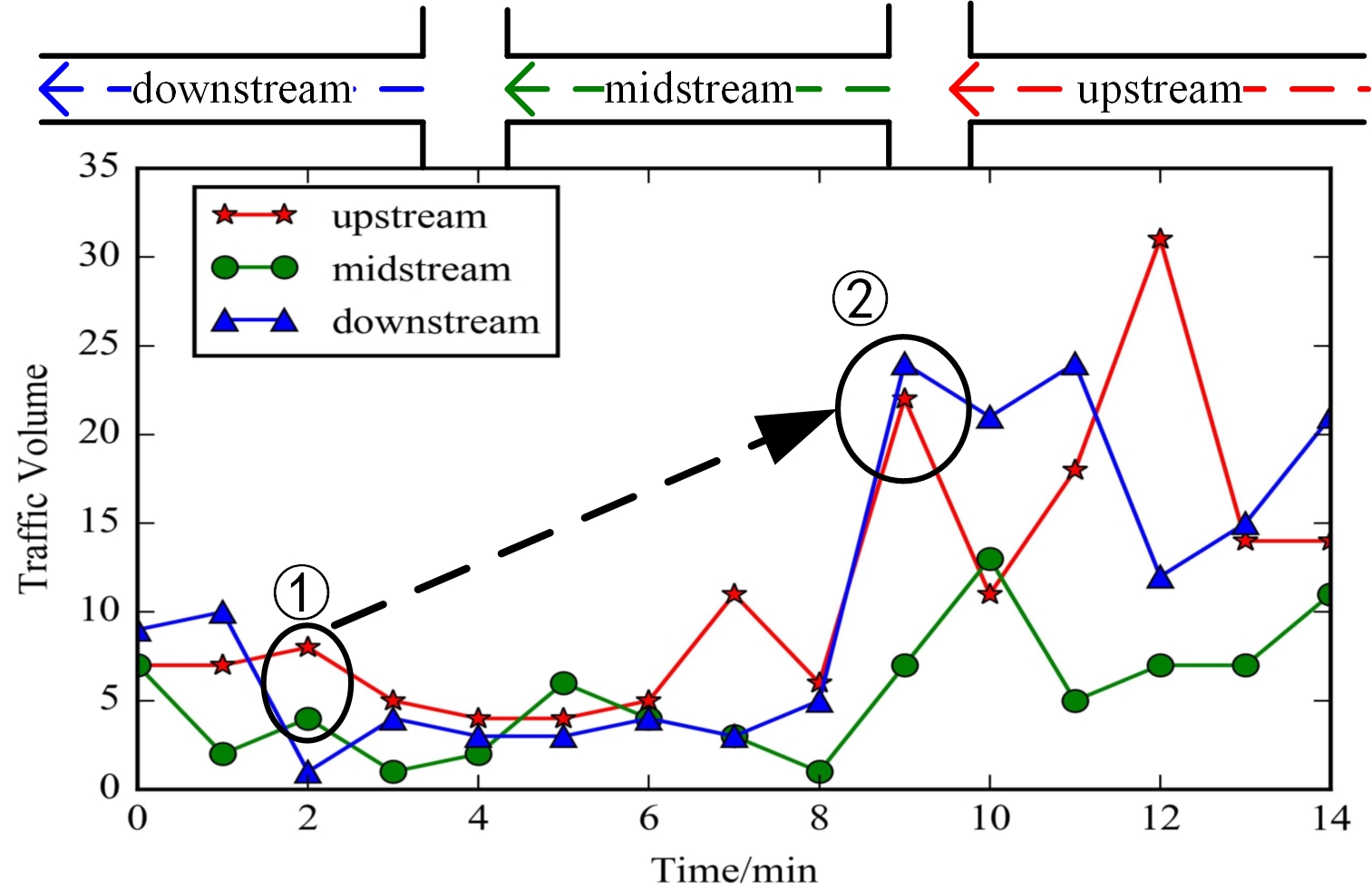}
	\caption{Spatial dependence is restricted by the topological structure of the road network. Due to the strong influence between adjacent roads, the short-term traffic flow similarity is changed from state \textcircled{1} to state \textcircled{2}.}
	\label{spatial dependence}
\end{figure}

%
%
%
%
\IEEEPARstart{W}{ith} the development of the Intelligent Traffic System, traffic forecasting has received more and more attention. It is a key part of an advanced traffic management system and is an important part of realizing traffic planning, traffic management, and traffic control. Traffic forecasting is a process of analyzing traffic conditions on urban roads, including flow, speed, and density, mining traffic patterns, and predicting the trends of traffic on roads. Traffic forecasting can not only provide a scientific basis for traffic managers to sense traffic congestions and limit vehicles in advance but also provide security for urban travelers to choose appropriate travel routes and improve travel efficiency \cite{Huang2005Dyanamic},\cite{Liu2004A},\cite{Yuan2012Synthesis}. However, traffic forecasting has always been a challenge task due to its complex spatial and temporal dependences:

(1) Spatial dependence. The change in traffic volume is dominated by the topological structure of the urban road network. The traffic status at upstream roads impact traffic status at downstream roads through the transfer effect, and the traffic status at downstream roads impact traffic status at upstream through the feedback effect \cite{Dong2012Spatial}. As shown in Figure \ref{spatial dependence}, due to the strong influence between adjacent roads, the short-term similarity is changed from state \textcircled{1} (the upstream road is similar to the midstream road) to state \textcircled{2} (the upstream road is similar to the downstream road).

\begin{figure}
	\centering
	\subfigure[]{\includegraphics[width=0.95\linewidth]{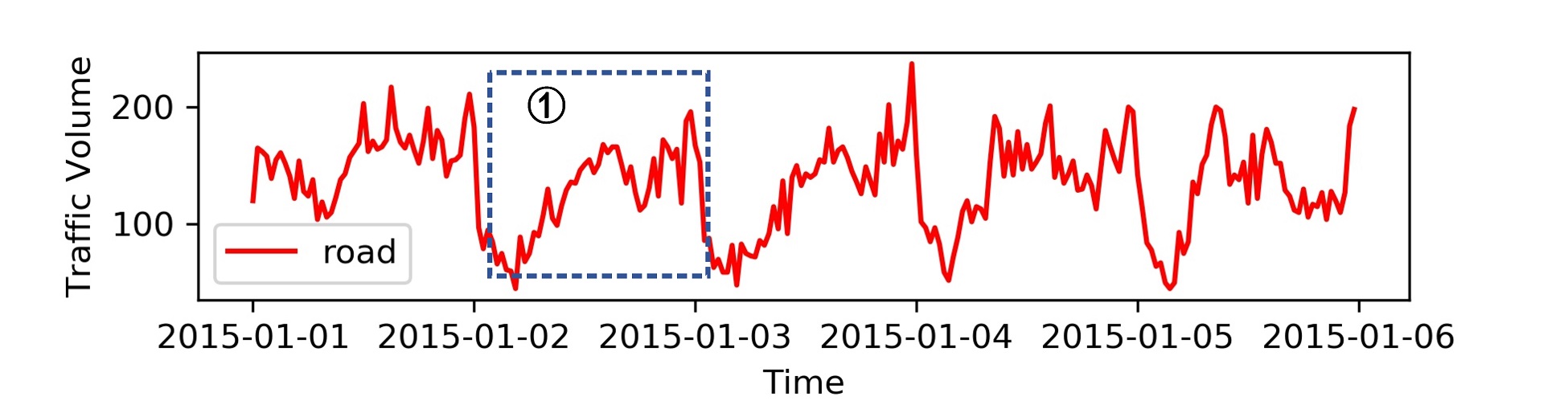}}
	\subfigure[]{\includegraphics[width=0.95\linewidth]{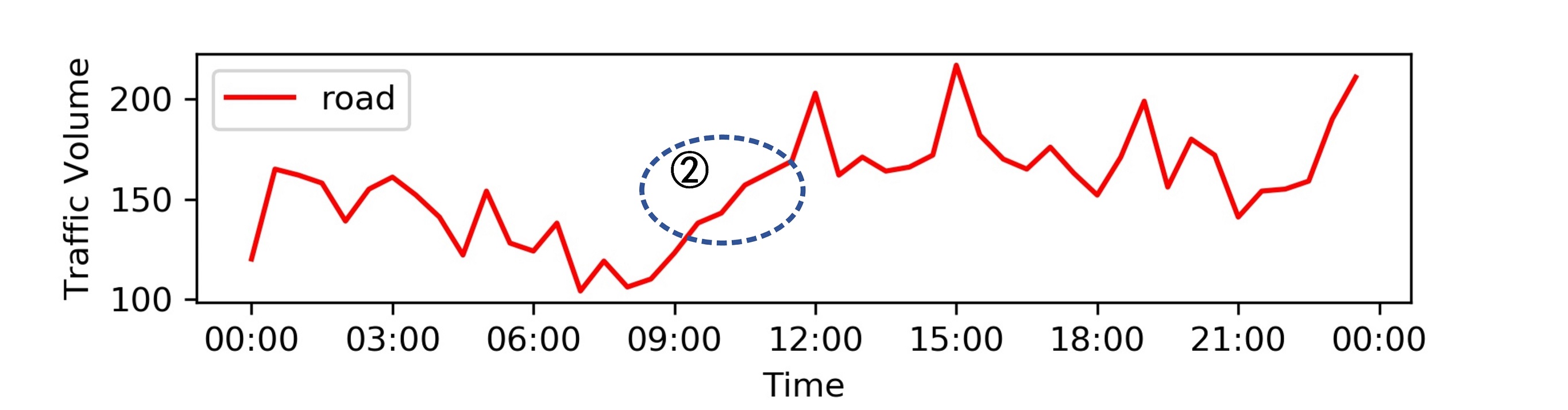}}
	\caption{(a) Periodicity. The traffic volume in the road changes periodically within one week. (b) Trend. The traffic volume in the road has tendency change within one day.}
	\label{temporal dependence}
\end{figure}

(2) Temporal dependence. The traffic volume changes dynamically over time and is mainly reflected in periodicity and trend. As shown in Figure \ref{temporal dependence}(a), the traffic volume on Road 1 shows a periodic change over a week. As shown in Figure \ref{temporal dependence}(b), the traffic volume in one day will also change over time; for example, the traffic volume will be affected by the traffic condition of the previous moment or even longer.

There are many existing traffic forecasting methods, some of which consider temporal dependence, including the Autoregressive Integrated Moving Average (ARIMA) model \cite{Ahmed1979ANALYSIS}, \cite{Hamed1995Short}, the Kalman filtering model \cite{Okutani1984Dynamic}, the support vector regression machine model \cite{Wu2004Travel},\cite{Yao2006Research}, the k-nearest neighbor model \cite{Zhang2009Short}, the Bayesian model \cite{Sun2006A}, and partial neural network model \cite{Huang2014Deep},\cite{Fu2017Using}. The above methods consider the dynamic change of traffic condition but ignore the spatial dependence, so that the change of traffic condition is not restricted by the road network and we cannot predict the state of traffic data accurately. To characterize the spatial features better, some studies \cite{Zhang2016Deep},\cite{Wu2016Short},\cite{Cao2017Interactive} introduce a convolution neural network for spatial modeling; however, a convolutional neural network is commonly used for Euclidean data \cite{Defferrard2016Convolutional} such as images, regular grids, and so on. Such models cannot work under the context of an urban road network with a complex topological structure so in essence they cannot describe the spatial dependence. 

To solve the above problems, we propose a new traffic forecasting method called the temporal graph convolutional network (T-GCN), which is used for traffic forecasting task based on urban road network. Our contributions are three-fold:

(1) We propose the T-GCN model by combining the graph convolutional network and gated recurrent unit. The graph convolutional network is used to capture the topological structure of the road network to model spatial dependence. The gated recurrent unit is used to capture the dynamic change of traffic data on the roads to model temporal dependence. The T-GCN model can also be applied to other spatio-temporal forecasting tasks.

(2) The forecasting results of the T-GCN model show a steady state under different prediction horizons, which indicates that the T-GCN model can not only achieve short-term prediction but can also be used for long-term traffic prediction tasks.

(3) We evaluate our approach using the taxi speed data of the Luohu District in Shenzhen and Los-loop datasets. The results show that our approach reduces the prediction error by approximately 1.5$\%$-57.8$\%$ compared to all baseline methods, which demonstrates that the T-GCN model has superiority in traffic forecasting.

The rest of the paper is organized as follows. Section II reviews relevant research about traffic forecasting. Section III introduces the details of our method. In section IV, we evaluate the predictive performance of the T-GCN by real-world traffic dataset, including design of model parameters, prediction results analysis, perturbation analysis, and model interpretation. We conclude the paper in Section V.

\section{Related Work}
Intelligent traffic system traffic forecasting is one of the major research issues today. The existing traffic forecasting methods can be divided into two categories: the model-driven approach and the data-driven approach. First, the model-driven approach mainly explains the instantaneous and steady-state relationships among traffic volume, speed, and density. Such methods require comprehensive and detailed system modeling based on prior knowledge. The representative methods contain the queuing theory model \cite{Xu2014Analysis}, the cell transmission model \cite{Wei2013Total}, the “traffic velocity” model\cite{Qi2011Traffic}, the microscopic fundamental diagram model\cite{Xu2013Impacts}, and so on. In reality, traffic data is influenced by many factors and it is difficult to obtain an accurate traffic model. The existing models cannot accurately describe the variations of traffic data in complex real-world environments. In addition, the construction of these models requires significant computing capability \cite{Vlahogianni2015Computational} and is easily constrained by traffic disturbances and sampling point spacing, etc.

Second, data-driven approaches infer the variation tendency based on statistical regularity of the data and is eventually used to predict and evaluate the traffic state \cite{Shan2013Urban}, \cite{Shen2011Short}. This type of method does not analyze the physical properties and dynamic behavior of the traffic system and has high flexibility. The earlier method includes the historical average model \cite{Liu2004A}, in which the average value of the traffic volume in historical periods is used as the prediction value. This method does not require any assumptions and the calculation is simple and fast but it cannot fit well with temporal features and the prediction precision is low. With the continuous deepening of research on traffic forecasting, a large number of methods with higher prediction precision have emerged, which can be mainly divided into a parametric model and nonparametric model \cite{Eleni2004Short},\cite{Van2012Short}.

The parametric model presupposes the regression function, the parameters are determined through processing the original data, and then realizing the traffic forecasting is based on the regression function. The time series model, the linear regression model \cite{Sun2004Interval},\cite{Dudek2016Pattern}, and the Kalman filtering model are common methods. The time series model fits the observed time series into a parametric model to predict future data. As early as 1976, Box and Jenkins \cite{Ahmed1979ANALYSIS} proposed the Autoregressive Integrate Moving Average Model (ARIMA), which is the most widely used time series model. In 1995, Hamed et al. \cite{Hamed1995Short} used the ARIMA model to predict the traffic volume in urban arterials. To improve the prediction precision of the model, different variants were produced, including Kohonen ARIMA \cite{Voort1996Combining}, subset ARIMA \cite{Lee1999Application}, seasonal ARIMA \cite{Fabian2003Modeling}, and so on. Lippi et al. \cite{Lippi2013Short} compared the support vector regression model with the seasonal ARIMA model and found that the SARIMA model has better results in traffic congestion. The linear regression model builds a regression function based on historical traffic data to predict traffic flow. In 2004, Sun et al. \cite{Sun2004Interval} solves the problem of interval forecasting using the local linear model, and obtain better result on the real-world traffic dataset. The Kalman filtering model predicts future traffic conditions based on the traffic state of the previous moment and the current moment. In 1984, Okutani et al. \cite{Okutani1984Dynamic} used the Kalman filtering theory to establish the traffic flow state prediction model. Subsequently, some studies \cite{Hinsbergen2012Localized},\cite{Ojeda2013Adaptive} used the Kalman filtering model to realize traffic prediction tasks.

The traditional parametric model has a simple algorithm and convenient calculation. However, these models depend on the assumption of stationary, cannot reflect the nonlinearity and uncertainty characteristics of traffic data, and cannot overcome the interference of random events such as traffic accidents. The nonparametric model solves these problems well and only requires enough historical data to learn the statistical regularity from traffic data automatically. The common nonparametric model includes: the k-nearest neighbor model \cite{Zhang2009Short}, the support vector regression model \cite{Wu2004Travel},\cite{Yao2006Research},\cite{Smola2004A}, the Fuzzy Logic model \cite{Yin2002Urban}, the Bayesian network model \cite{Sun2006A}, the neural network model, and so on.

In recent years, with the rapid development of deep learning \cite{Silver2016Mastering},\cite{Silver2017Mastering},\cite{Morav2017DeepStack}, the deep neural network models have received attention because they can capture the dynamic characteristics of traffic data well and achieve the best results at present. According to whether or not spatial dependence is considered, models can be divided into two categories. Some methods consider temporal dependence only, e.g., Park et al. \cite{Park2010Forecasting} used Feed Forward NN to implement traffic flow prediction tasks. Huang et al. \cite{Huang2014Deep} proposed a network architecture consisting of a deep belief network (DBN) and a regression model and verified that the network can capture random features from traffic data on multiple datasets and this model improved prediction accuracy in traffic forecasting. In addition, since the recurrent neural network (RNN) and its variants have long short-term memory (LSTM) and the gated recurrent unit (GRU) can effectively use the self-circulation mechanism, they can learn temporal dependence well and achieve better prediction results \cite{Fu2017Using},\cite{J2002FREEWAY}.

These models take the temporal feature into account but ignore the spatial dependence, so that the change of traffic data is not constrained by the urban road network and thus they cannot accurately predict the traffic state on the road. Making full use of the spatial and temporal dependence is the key to solving traffic forecasting problems. To better characterize spatial features, many studies had made improvements on this basis. Lv et al. \cite{Lv2015Traffic} proposed a SAE model to capture the spatio-temporal feature from traffic data and realize short-term traffic flow predictions. Zhang et al. \cite{Zhang2016Deep} proposed a deep learning model called ST-ResNet, which designed residual convolutional networks for each attribute based on the temporal closeness, period, and trend of crowd flows, and then three networks and external factors were dynamically aggregated to predict the inflow and outflow of crowds in each region of a city. Wu et al. \cite{Wu2016Short} designed a feature fusion architecture for short-term prediction by combining CNN and LSTM. A 1-dimensional CNN was used to capture spatial dependence and two LSTMs were used to mine the short-term variability and periodicity of traffic flow. Cao et al. \cite{Cao2017Interactive} proposed an end-to-end model called ITRCN, which converted interactive network traffic into images and used CNN to capture interactive functions of traffic, used GRU to extract temporal features, and proved that the prediction error of this model is 14.3\% and 13.0\% higher than that of GRU and CNN, respectively. Ke et al. \cite{Ke2017Short} proposed a new deep learning method called the fusion convolutional long short-term memory network (FCL-Net), taking into account spatial dependence, temporal dependence, and exogenous dependence for short-term passenger demand forecasting. Yu et al. \cite{Yu2017spatio-temporal} used the Deep Convolutional Neural Network(DCNN) to capture spatial dependence, used LSTM to capture temporal dynamics, and demonstrated the superiority of the SRCN model through experiments on Beijing traffic network data.

Although the above methods introduced the CNN to model spatial dependence and made great progress in traffic forecasting tasks, the CNN is essentially suitable for Euclidean space, such as images, regular grids, etc., and has limitations on traffic networks with a complex topological structure, and thus cannot essentially characterize the spatial dependence. Therefore, this type of method also has certain defects. In recent years, with the development of the graph convolutional network model \cite{Kipf2016Semi}, which can be used to capture structural feature of graph network, provides a good solution for the above problem. Li et al. \cite{Li2017Graph} proposed a DCRNN model, which captures the spatial feature through random walks on graphs, and the temporal feature through encoder-decoder architecture.

Based on this background, in this paper we propose a new neural network approach that can capture the complex temporal and spatial features from traffic data, and can then be used for traffic forecasting tasks based on an urban road network.

\section{Methodology}
\subsection{Problem Definition}
In this paper, the goal of the traffic forecasting is to predict the traffic information in a certain period of time based on the historical traffic information on the roads. In our method, the traffic information is a general concept which can be traffic speed, traffic flow, and traffic density. Without loss of generality, we use traffic speed as a example of traffic information in experiment section.

Definition 1: road network G. We use an unweighted graph $G=(V,E)$ to describe the topological structure of the road network, and we treat each road as a node, where V is a set of road nodes, $V=\left\{v_1,v_2,\cdots,v_N\right\}$, N is the number of the nodes, and E is a set of edges. The adjacency matrix A is used to represent the connection between roads, $A\in R^{N\times N}$. The adjacency matrix contains only elements of 0 and 1. The element is 0 if there is no link between roads and 1 denotes there is a link.

Definition 2: feature matrix $X^{N\times P}$. We regard the traffic information on the road network as the attribute feature of the node in the network, expressed as $X\in R^{N\times P}$, where P represents the number of node attribute features (the length of the historical time series) and $X_{t}\in R^{N\times i}$ is used to represent the speed on each road at time i. Again, the node attribute features can be any traffic information such as traffic speed, traffic flow, and traffic density.

Thus, the problem of spatio-temporal traffic forecasting can be considered as learning the mapping function f on the premise of road network topology G and feature matrix X and then calculating the traffic information in the next T moments, as shown in equation 1:
\begin{equation}
\left[X_{t+1},\cdots,X_{t+T}\right] = f\left(G;\left(X_{t-n},\cdots,X_{t-1},X_{t}\right)\right) 
\end{equation}

where n is the length of historical time series and T is the length of the time series needed to be predicted.
\subsection{Overview}
\begin{figure}
	\centering
	\includegraphics[width=0.9\linewidth]{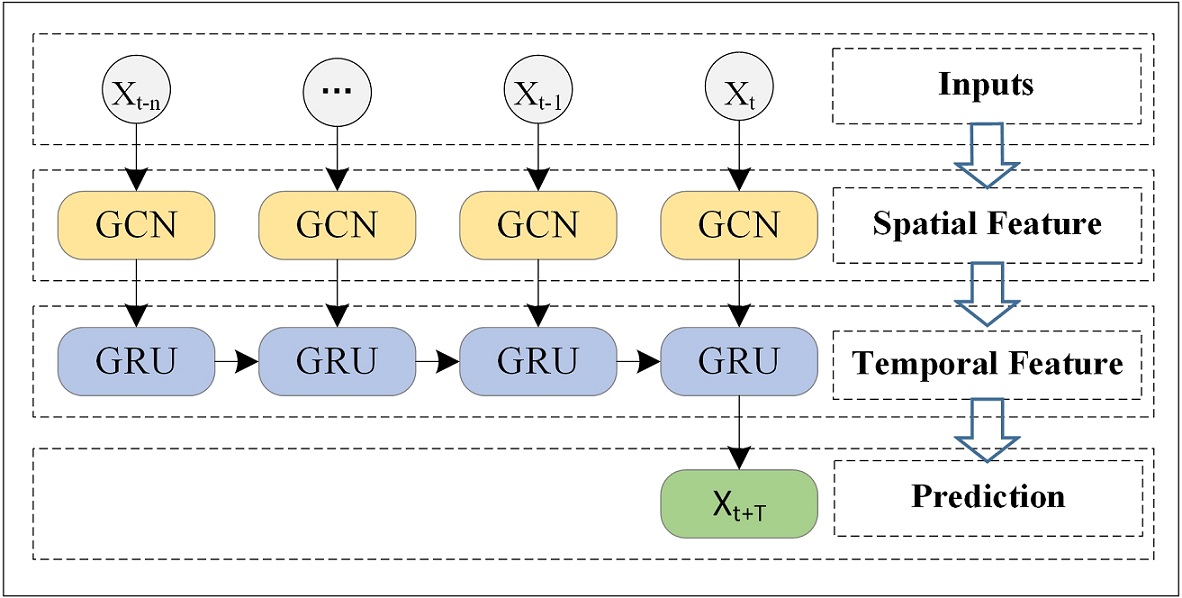}
	\caption{Overview. We take the historical traffic information as input and obtain the finally prediction result through the Graph Convolution Network and the Gated Recurrent Units model.}
	\label{overview}
\end{figure}

In this section, we describe how to use the T-GCN model to realize the traffic forecasting task based on the urban roads. Specifically, the T-GCN model consists of two parts: the graph convolutional network and the gated recurrent unit. As shown in Figure \ref{overview}, we first use the historical n time series data as input and the graph convolution network is used to capture topological structure of urban road network to obtain the spatial feature. Second, the obtained time series with spatial features are input into the gated recurrent unit model and the dynamic change is obtained by information transmission between the units, to capture temporal feature. Finally, we get results through the fully connected layer.

\subsection{Methodology}
\subsubsection{Spatial Dependence Modeling}
\begin{figure}
	\centering
	\includegraphics[width=0.8\linewidth]{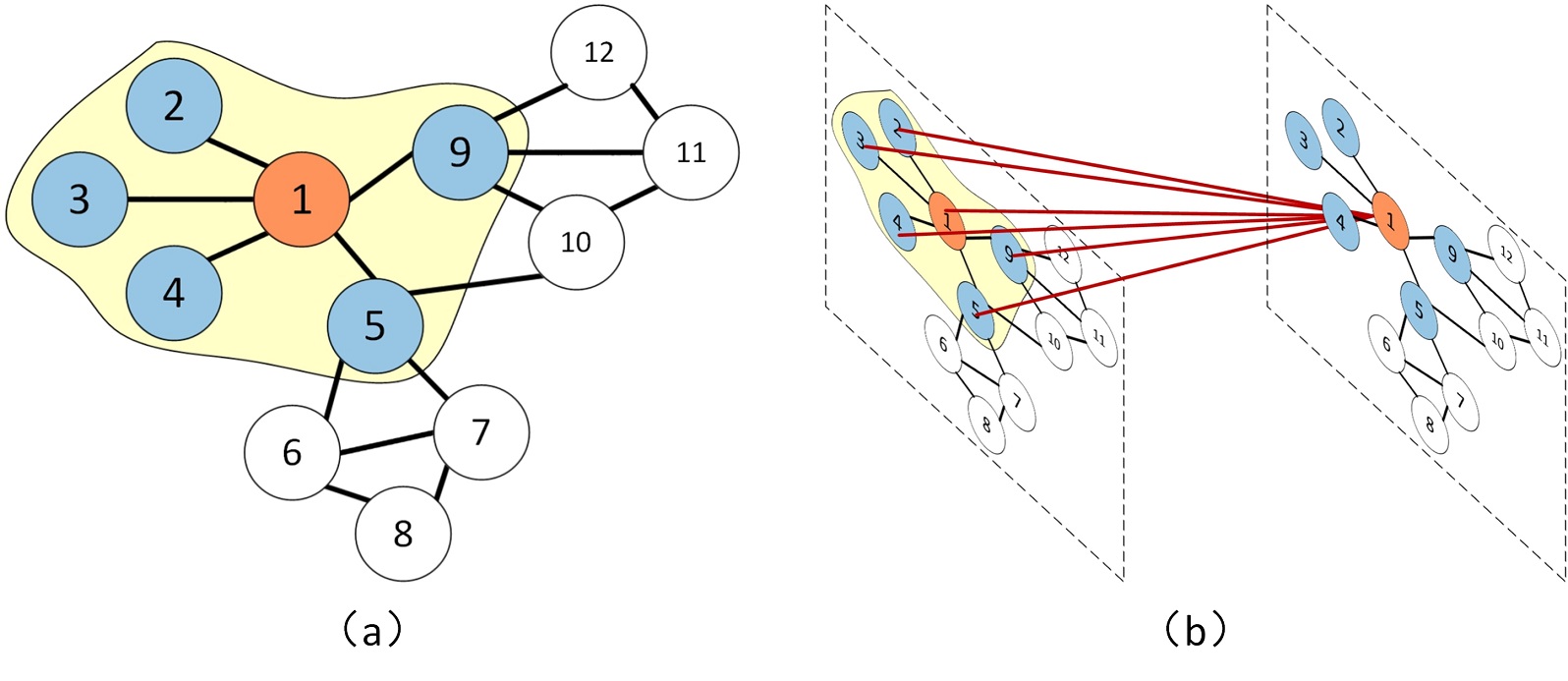}
	\caption{Assuming that node 1 is a central road. (a) The blue nodes indicate the roads connected to the central road. (b) We obtain the spatial feature by obtaining the topological relationship between the road 1 and the surrounding roads.}
	\label{spatial modeling}
\end{figure}

Acquiring the complex spatial dependence is a key problem in traffic forecasting. The traditional convolutional neural network (CNN) can obtain local spatial features, but it can only be used in Euclidean space, such as images, a regular grid, etc. An urban road network is in the form of graph rather than two-dimensional grid, which means the CNN model cannot reflect the complex topological structure of the urban road network and thus cannot accurately capture spatial dependence. Recently, generalizing the CNN to the graph convolutional network (GCN), which can handle arbitrary graph-structured data, has received widespread attention. The GCN model has been successfully used in many applications, including document classification \cite{Defferrard2016Convolutional}, unsupervised learning \cite{Kipf2016Semi} and image classification \cite{Bruna2014Spectral}. The GCN model constructs a filter in the Fourier domain, the filter acts on the nodes of graph and its first-order neighborhood to capture spatial features between the nodes, and then the GCN model can be built by stacking multiple convolutional layers. As shown in Figure \ref{spatial modeling}, assuming that node 1 is the central road, the GCN model can obtain the topological relationship between the central road and its surrounding roads, encode the topological structure of the road network and the attributes on the roads, and then obtain spatial dependence. In summary, we use the GCN model \cite{Bruna2014Spectral} to learn spatial features from traffic data. A 2-layer GCN model can be expressed as:
\begin{equation}
f\left(X,A\right) = \sigma\left(\widehat{A}Relu\left(\widehat{A}XW_{0}\right)W_{1}\right)
\end{equation}
where X represents the feature matrix, A represents the adjacency matrix, $\widehat{A} = \widetilde{D}^{-\frac{1}{2}}\widetilde{A}\widetilde{D}^{-\frac{1}{2}}$denotes preprocessing step, $\widetilde{A}=A+I_{N}$ is a matrix with self-connection structure, $\widetilde{D}$ is a degree matrix, $\widetilde{D}=\sum_{j}\widetilde{A}_{ij}$. $W_{0}$ and $W_{1}$ represent the weight matrix in the first and second layer, and $\sigma(\cdot)$, $Relu()$ represent the activation function.

\subsubsection{Temporal Dependence Modeling}
\begin{figure}
	\centering
	\includegraphics[width=3.5in,height=1.2in]{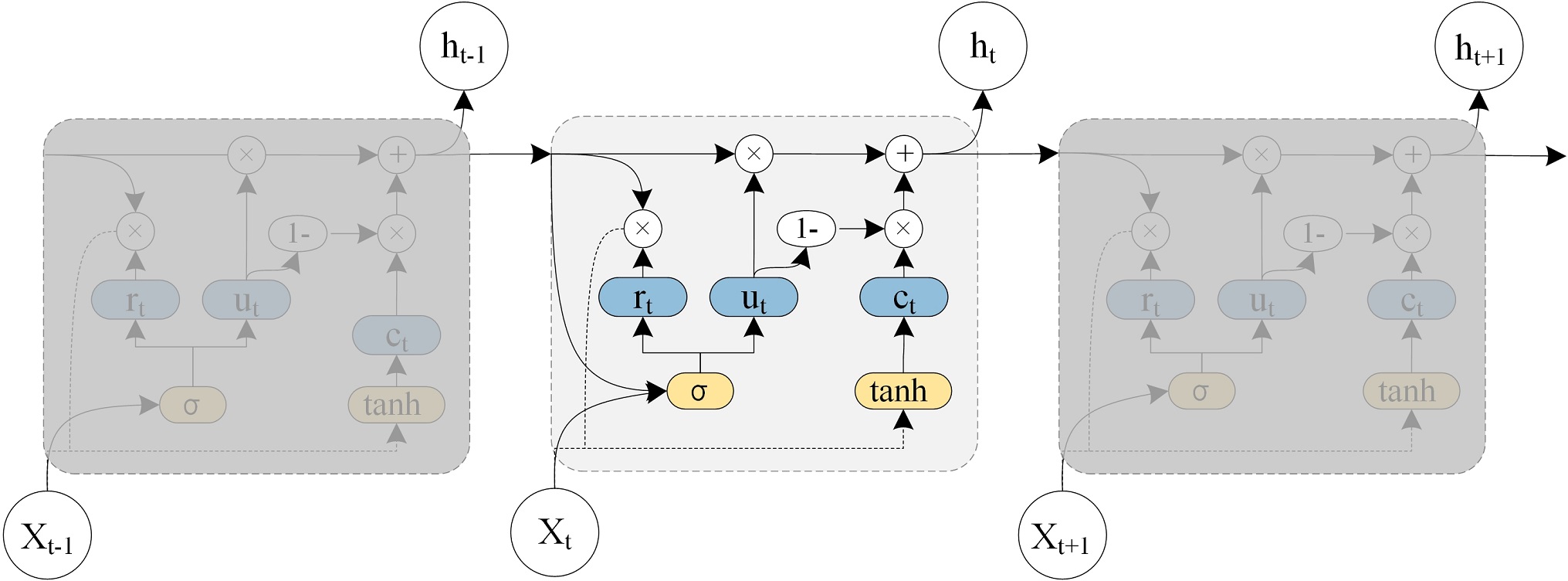}
	\caption{The architecture of the Gated Recurrent Unit model.}
	\label{temporal modeling}
\end{figure}

Acquiring the temporal dependence is another key problem in traffic forecasting. At present, the most widely used neural network model for processing sequence data is the recurrent neural network (RNN). However, due to defects such as gradient disappearance and gradient explosion, the traditional recurrent neural network has limitations for long-term prediction \cite{Bengio2002Learning}. The LSTM model \cite{Sepp1997Long} and the GRU model \cite{Cho2014On} are variants of the recurrent neural network and have been proven to solve the above problems. The basic principles of the LSTM and GRU are roughly the same \cite{Chung2014Empirical}. they all use gated mechanism to memorize as much long-term information as possible and are equally effective for various tasks. However, due to its complex structure, LSTM has a longer training time while the GRU model has a relatively simple structure, fewer parameters, and faster training ability. Therefore, we chose the GRU model to obtain temporal dependence from the traffic data. As shown in Figure \ref{temporal modeling}, $h_{t-1}$ denotes the hidden state at time t-1; $x_{t}$ denotes the traffic information at time t; $r_{t}$ is the reset gate, which is used to control the degree of ignoring the status information at the previous moment; $u_{t}$ is the update gate, which is used to control the degree of to which the status information at the previous time is brought into the current status; $c_{t}$ is the memory content stored at time t; and $h_{t}$ is output state at time t. The GRU obtains the traffic status at time t by taking the hidden status at time t-1 and the current traffic information as inputs. While capturing the traffic information at the current moment, the model still retains the changing trend of historical traffic information and has the ability to capture temporal dependence.

\subsubsection{Temporal Graph Convolutional Network}
\begin{figure}
	\centering
	\includegraphics[width=0.9\linewidth]{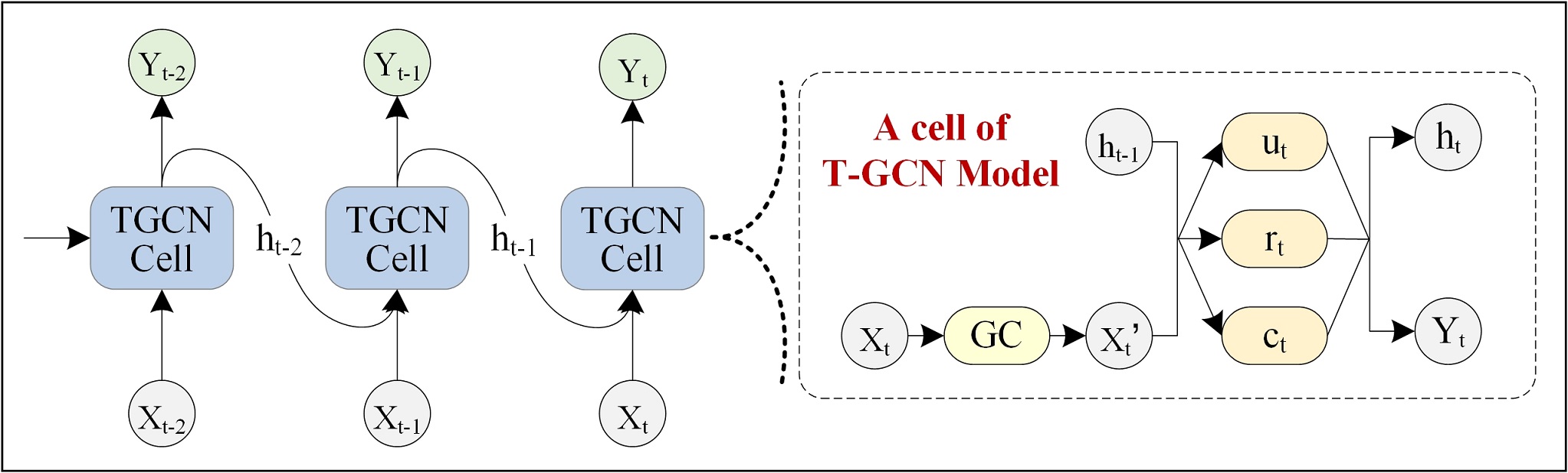}
	\caption{The overall process of spatio-temporal prediction. The right part represents the specific architecture of a T-GCN unit, and GC represents graph convolution.}
	\label{T-gcn}
\end{figure}

To capture the spatial and temporal dependences from traffic data at the same time, we propose a temporal graph convolutional network model (T-GCN) based on a graph convolutional network and gated recurrent units. As shown in Figure \ref{T-gcn}, the left side is the process of spatio-temporal traffic prediction, the right side shows the specific structure of a T-GCN cell, $h_{t-1}$ denotes the output at time t-1, GC is graph convolution process, and $u_{t}$, $r_{t}$ are update gate and reset gate at time t, and $h_{t}$ denotes the output at time t.
The specific calculation process is shown below. $f\left(A,X_{t}\right)$ represents the graph convolution process and is defined in equation 2. W and b represent the weights and deviations in the training process.
\begin{equation}
u_{t}=\sigma(W_{u}∙\left[f(A,X_{t}),h_{t-1}\right]+b_{u})
\end{equation}
\begin{equation}
r_{t}=\sigma(W_{r}∙\left[f(A,X_{t}),h_{t-1}\right]+b_{r})
\end{equation}
\begin{equation}
c_{t}=tanh(W_{c}\left[f(A,X_{t}),(r_{t}*h_{t-1})\right]+b_{c})
\end{equation}
\begin{equation}
h_{t}=u_{t}*h_{t-1}+(1-u_{t})*c_{t}
\end{equation}

In summary, the T-GCN model can deal with the complex spatial dependence and temporal dynamics. On one hand, the graph convolutional network is used to capture the topological structure of the urban road network to obtain the spatial dependence. On the other hand, the gated recurrent unit is used to capture the dynamic variation of traffic information on the roads to obtain the temporal dependence and eventually realize traffic prediction tasks.

\subsubsection{Loss Function}
In the training process, the goal is to minimize the error between the real traffic speed on the roads and the predicted value. We use $Y_{t}$ and $\widehat{Y_{t}}$ to denote the real traffic speed and the predicted speed, respectively. The loss function of the T-GCN model is shown in equation 7. The first term is used to minimize the error between the real traffic speed and the prediction. The second term $L_{reg}$ is an L2 regularization term that helps to avoid an over fitting problem and $\lambda$ is a hyperparameter.
\begin{equation}
loss=\parallel Y_{t}-\widehat{Y_{t}}\parallel+\lambda L_{reg}
\end{equation}

\section{Experiments}

\subsection{Data Description}
In this section, we evaluate the prediction performance of the T-GCN model on two real-world datasets: SZ-taxi dataset and Los-loop set. Since these two datesets are all related to traffic speed. Without loss of generality, we use traffic speed as traffic information in experiment section.

(1) SZ-taxi. This dataset was the taxi trajectory of Shenzhen from Jan. 1 to Jan. 31, 2015. We selected 156 major roads of Luohu District as the study area. The experimental data mainly includes two parts. One is an 156*156 adjacency matrix, which describes the spatial relationship between roads. Each row represents one road and the values in the matrix represent the connectivity between the roads. Another one is a feature matrix, which describes the speed changes over time on each road. Each row represents one road; each column is the traffic speed on the roads in different time periods. We aggregate the traffic speed on each road every 15 minutes.

(2) Los-loop. This dataset was collected in the highway of Los Angeles County in real time by loop detectors. We selected 207 sensors and it’s traffic speed from Mar.1 to Mar.7, 2012. We aggregated the traffic speed every 5 minutes. Similarity, the data concludes an adjacency matrix and a feature matrix. The adjacency matrix is calculated by the distance between sensors in the traffic networks. Since the Los-loop dataset contained some missing data, we used the linear interpolation method to fill missing values.

In the experiments,the input data was normalized to the interval [0,1]. In addition, 80$\%$ of the data was used as the training set and the remaining 20$\%$ was used as the testing set. We predicted the traffic speed of the next 15 minutes, 30 minutes, 45 minutes and 60 minutes.

\subsection{Evaluation Metrics}
To evaluate the prediction performance of the T-GCN model, we use five metrics to evaluate the difference between the real traffic information $Y_t$ and the prediction $\widehat{Y_{t}}$, including:

(1) Root Mean Squared Error (RMSE):
\begin{equation}
RMSE=\sqrt{\frac{1}{n}\sum_{i=1}^{n}(Y_{t}-\widehat{Y_{t}})^{2}}
\end{equation}

(2) Mean Absolute Error (MAE):
\begin{equation}
MAE=\frac{1}{n}\sum_{i=1}^{n}\left|Y_{t}-\widehat{Y_{t}}\right|
\end{equation}

(3) Accuracy:
\begin{equation}
Accuracy=1-\frac{\parallel Y-\widehat{Y}\parallel_{F}}{\parallel Y\parallel_{F}}
\end{equation}

(4) Coefficient of Determination (R2):
\begin{equation}
R^{2}=1-\frac{\sum_{i=1}(Y_{t}-\widehat{Y_{t}})^{2}}{\sum_{i=1}(Y_{t}-\bar{Y})^{2}}
\end{equation}

(5) Explained Variance Score (Var):
\begin{equation}
var=1-\frac{Var\left\{Y-\widehat{Y}\right\}}{Var\left\{Y\right\}}
\end{equation}

Specifically, RMSE and MAE are used to measure the prediction error: the smaller the value is, the better the prediction effect is. Accuracy is used to detect the prediction precision: the lager the value is, the better the prediction effect is. $R^{2}$ and Var calculate the correlation coefficient, which measures the ability of the prediction result to represent the actual data: the larger the value is, the better the prediction effect is.

\subsection{Model Parameters Designing}
(1) Hyperparameter

The hyperparameters of the T-GCN model mainly include: learning rate, batch size, training epoch, and the number of hidden layers. In the experiment, we manually adjust and set the learning rate to 0.001, the batch size to 64, and the training epoch to 3000.

The number of hidden units is a very important parameter of the T-GCN model, as different hidden units may greatly affect the prediction precision. To choose the best value, we experiment with different hidden units and select the optimal value by comparing the predictions.
\begin{figure*}
	\centering
	\subfigure[]{
	    \includegraphics[width=0.4\linewidth]{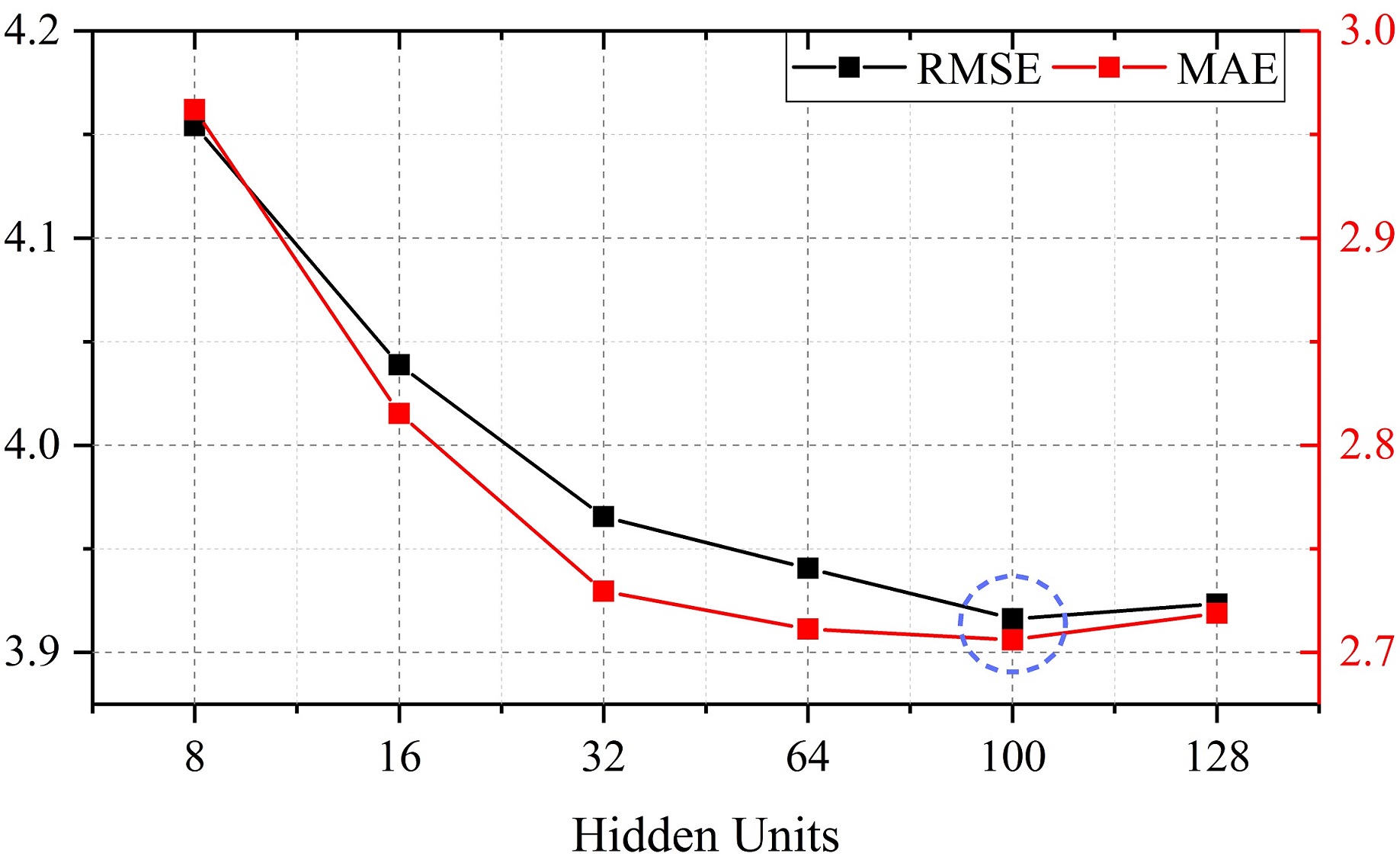}}
	    \label{hidden_rmse} 
	\subfigure[]{		
		\includegraphics[width=0.4\linewidth]{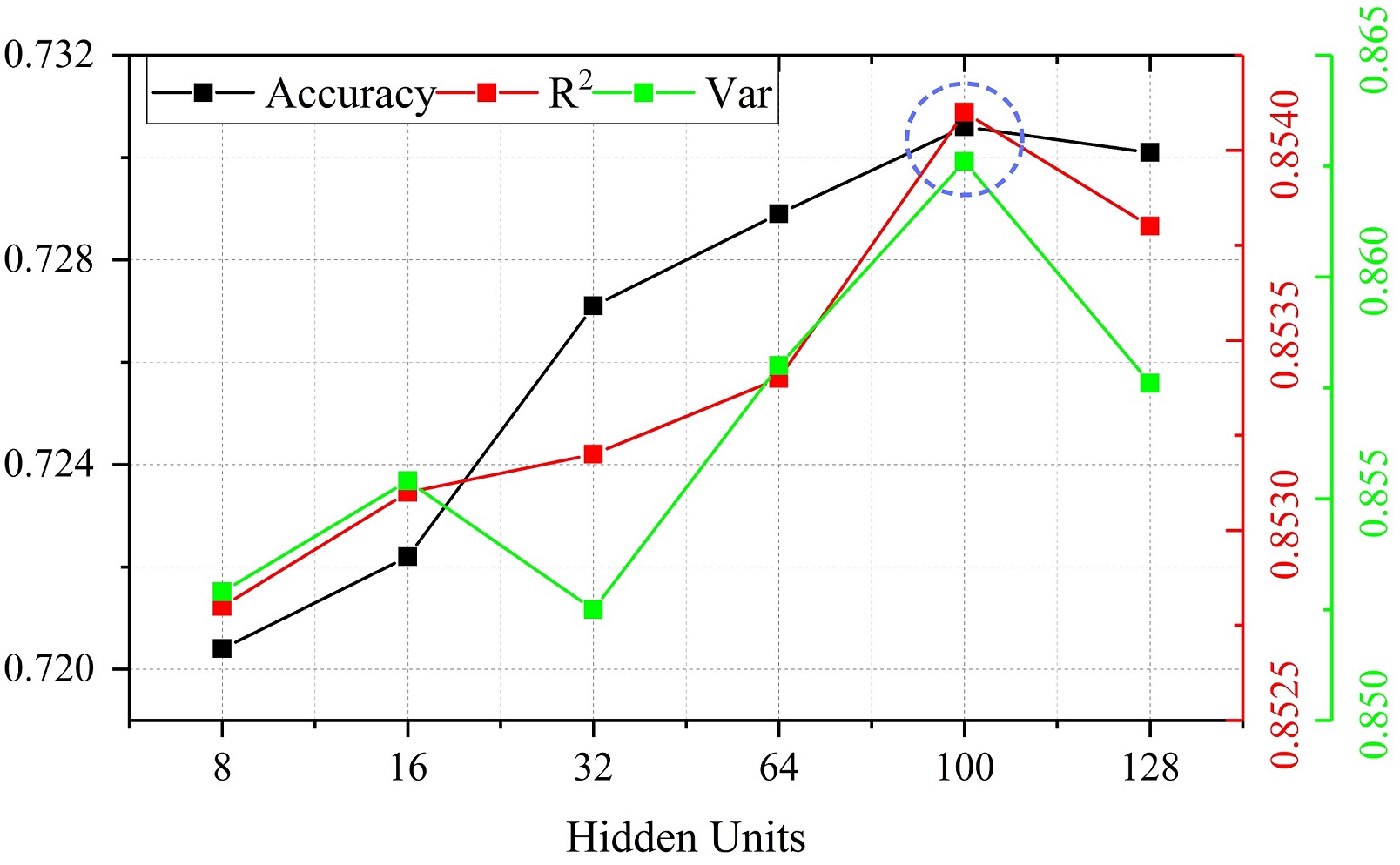}}
		\label{Accuracy} 
	\subfigure[]{		
		\includegraphics[width=0.4\linewidth]{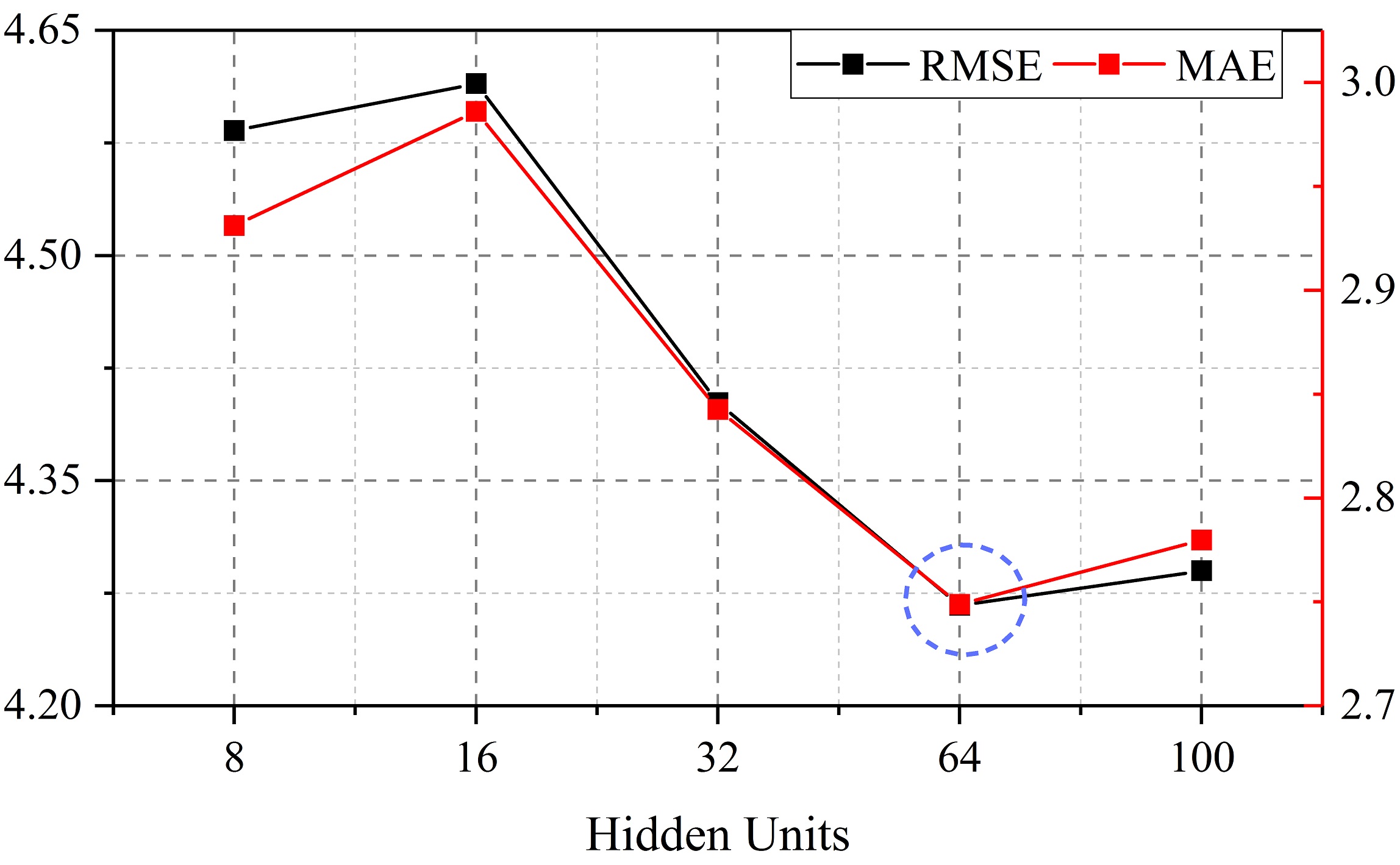}}
		\label{Los-hidden-error} 
	\subfigure[]{
		\includegraphics[width=0.4\linewidth]{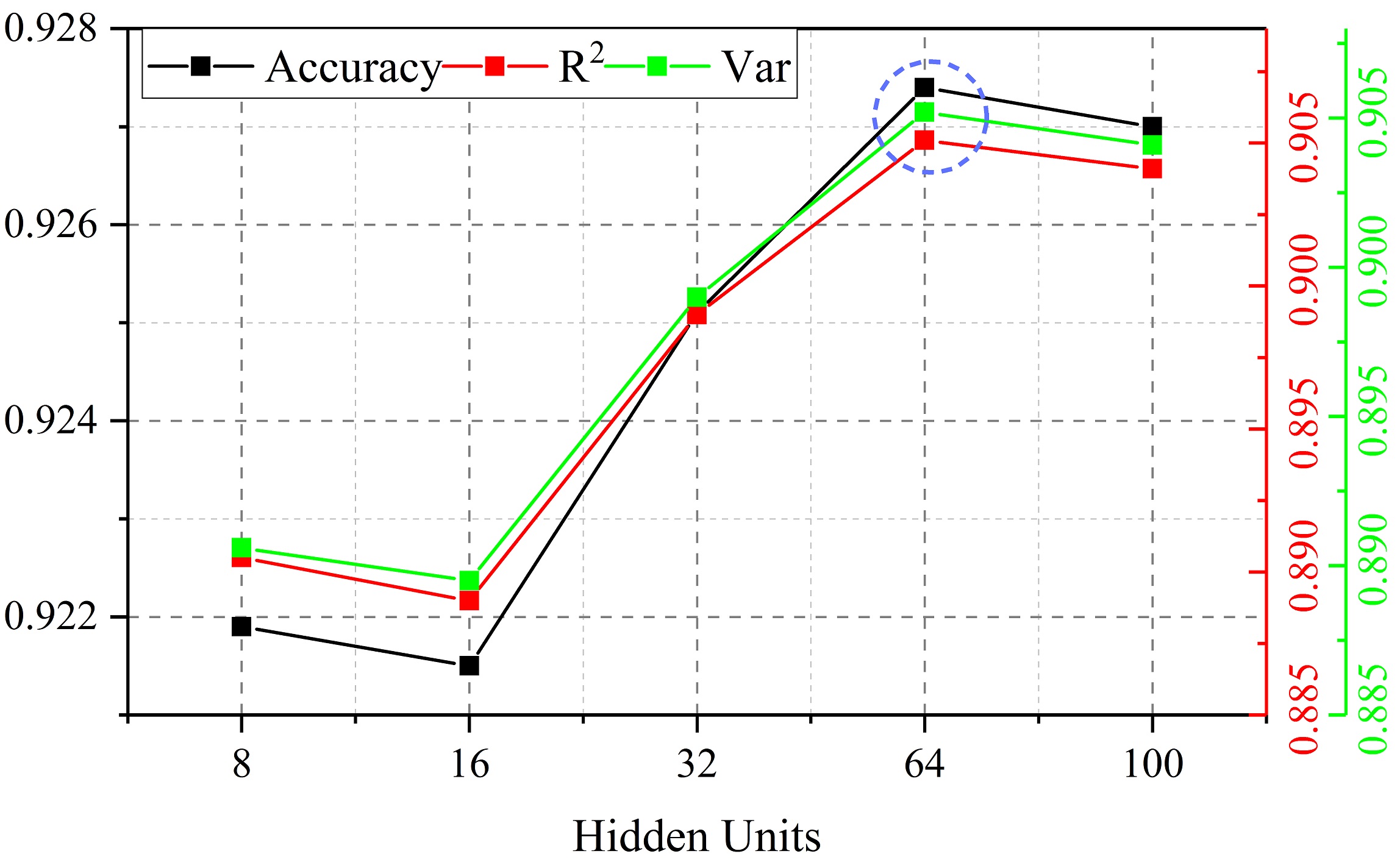}}
		\label{Los-Accuracy}
	\caption{Comparison of predicted performance under different hidden units. (a) Changes in RMSE and MAE based on SZ-taxi. (b) Changes in Accuracy, $R^{2}$ and Var based on SZ-taxi. (c) Changes in RMSE and MAE based on Los-loop. (d) Changes in Accuracy, $R^{2}$ and Var based on Los-loop.}
	\label{hidden}
\end{figure*}

In our experiment, for the SZ-taxi dataset, we choose the number of hidden units from [8, 16, 32, 64, 100, 128] and analyze the change of prediction precision. As shown in \ref{hidden}, the horizontal axis represents the number of hidden units and the vertical axis represents the change of different metrics. Figure \ref{hidden}(a) shows the results of RMSE and MAE for different hidden units. It can be seen that the error is the smallest when the number is 100. Figure \ref{hidden}(b) shows the variation of Accuracy, $R^{2}$, and Var for different hidden units. Similarly, when the number is 100, the results reach a maximum. In summary, the prediction results are best when the number is set to 100. When increasing the number of hidden units, the prediction precision first increases and then decreases. This is mainly because when the hidden unit is larger than a certain degree, the model complexity and the computational difficulty are greatly increased and as a result, the prediction precision will be reduced. Therefore, we set the number of hidden units to 100 in all experiments.

In the same way, the results of Los-loop are shown in \ref{hidden}(c) and \ref{hidden}(d), it can be seen that when the number of hidden units is 64, the prediction precision is highest, and the prediction error is lowest.

(2) Training

For input layer, the training dataset (80$\%$ of the overall data) is taken as input in the training process and the remaining data is used as input in the testing process. The T-GCM model is trained using the Adam optimizer.


\subsection{Experimental Results}

We compare the performance of the T-GCN model with the following baseline methods:

(1) History Average model (HA) \cite{Liu2004A}, which uses the average traffic information in the historical periods as the prediction.

(2) Autoregressive Integrated Moving Average model (ARIMA) \cite{Ahmed1979ANALYSIS}, which fits the observed time series into a parametric model to predict future traffic data.

(3) Support Vector Regression model (SVR) \cite{Smola2004A}, which uses historical data to train the model and obtains the relationship between the input and output, and then predicts by giving the future traffic data. We use the linear kernel and the penalty term is 0.001.

(4) Graph Convolutional Network model (GCN) \cite{Kipf2016Semi}: see 3.2.1 for details.

(5) Gated Recurrent Unit model (GRU) \cite{Cho2014On}: see 3.2.2 for details.

Table \ref{table} shows the T-GCN model and other baseline methods for 15 minutes, 30 minutes, 45 minutes and 60 minutes on SZ-taxi and Los-loop datasets. $\ast$ means that the values are too smal to be negligible, indicating that the model's prediction effect is poor. It can be seen that the T-GCN model obtains the best prediction performance under all evaluation metrics for all prediction horizons, proving the effectiveness of the T-GCN model in spatio-temporal traffic forecasting.

\begin{table*}
	\caption{The prediction results of the T-GCN model and other baseline methods on SZ-taxi and Los-loop datasets.}
	\centering
	\resizebox{160mm}{40mm}{
	\renewcommand{\arraystretch}{1.3}
	\begin{tabular}{c|c|cccccc|cccccc}
		\hline
		\multirow{2}{*}{T}&
		\multirow{2}{*}{Metric}&
		\multicolumn{6}{c|}{SZ-taxi}&
		\multicolumn{6}{c}{Los-loop} \\
		\cline{3-14}
		&&HA&ARIMA&SVR&GCN&GRU&T-GCN&HA&ARIMA&SVR&GCN&GRU&T-GCN\\
		\hline\hline
		\multirow{5}*{15min}
		&$RMSE$&7.9198&8.2151&7.5368&9.2717&4.0483&3.9162&7.4427&10.0439&6.0084&7.7922&5.2182&5.1264\\
		&$MA$E&5.4969&6.2192&4.9269&7.2606&2.6814&2.7061&4.0145&7.6832&3.7285&5.3525&3.0602&3.1802\\
		&$Accuracy$&0.6807&0.4278&0.6961&0.6433&0.7178&0.7306&0.8733&0.8275&	0.8977&0.8673&0.9109&0.9127\\
		&$R^{2}$&0.7914&0.0842&0.8111&0.6147&0.8498&0.8541&0.7121&$\ast$&0.8123&0.6843&0.8576&0.8634\\
		&$var$&0.7914&$\ast$&0.8121&0.6147&0.8499&0.8626&0.7121&$\ast$&0.8146&0.6844&0.8577&0.8634\\
		\hline
		\multirow{5}*{30min}
		&$RMS$E&7.9198&8.2123&7.4747&9.3450&4.0769&3.9617&7.4427&9.3450&6.9588&8.3353&6.2802&6.0598\\
		&$MAE$&5.4969&6.2144&4.9819&7.3211&2.7009&2.7452&4.0145&7.6891&3.7248&5.6118&3.6505&3.7466\\
		&$Accuracy$&0.6807&0.4281&0.6987&0.6405&0.7158&0.7275&0.8733&0.8275&0.8815&0.8581&0.8931&0.8968\\
		&$R^{2}$&0.7914&0.0834&0.8142&0.6086&0.8477&0.8523&0.7121&$\ast$&0.7492&0.6402&0.7957&0.8098\\
		&$var$&0.7914&$\ast$&0.8144&0.6086&0.8477&0.8523&0.7121&$\ast$&0.7523&0.6404&0.7958&0.8100\\
		\hline
		\multirow{5}*{45min}
		&$RMSE$&7.9198&8.2132&7.4755&9.4023&4.1002&3.9950&7.4427&10.0508&7.7504&8.8036&7.0343&6.7065\\
		&$MAE$&5.4969&6.2154&5.0332&7.3704&2.7207&2.7666&4.0145&7.6924&4.1288&5.9534&4.0915&4.1158\\
		&$Accuracy$&0.6807&0.4280&0.6986&0.6383&0.7142&0.7252&0.8733&0.8273&0.8680&0.8500&0.8801&0.8857\\
		&$R^{2}$&0.7914&0.0837&0.8141&0.6038&0.8460&0.8509&0.7121&$\ast$&0.6899&0.5999&0.7446&0.7679\\
		&$var$&0.7914&$\ast$&0.8142&0.6039&0.8459&0.8509&0.7121&$\ast$&0.6947&0.6001&0.7451&0.7684\\	\hline
		\multirow{5}*{60min}
		&$RMSE$&7.9198&8.2063&7.4883&9.4504&4.1241&4.0141&7.4427&10.0538&8.4388&9.2657&7.6621&7.2677\\
		&$MAE$&5.4969&6.2118&5.0714&7.4120&2.7431&2.7889&4.0145&7.6952&4.5036&6.2892&4.5186&4.6021\\
		&$Accuracy$&0.6807&0.4282&0.6981&0.6365&0.7125&0.7238&0.8733&0.8273&0.8562&0.8421&0.8694&0.8762\\
		&$R^{2}$&0.7914&0.0825&0.8135&0.5998&0.8442&0.8503&0.7121&$\ast$&0.6336&0.5583&0.6980&0.7283\\
		&$var$&0.7914&$\ast$&0.8136&0.5999&0.8321&0.8504&0.7121&$\ast$&0.5593&0.5593&0.6984&0.7290\\
		\hline
	\end{tabular}}
	\label{table}
\end{table*}

(1) High prediction precision. We can find that the neural network-based methods, including the T-GCN model, the GRU model, which emphasize the importance of modeling the temporal feature, generally have better prediction precision than other baselines, such as the HA model, the ARIMA model and the SVR model. For example, for the 15-min traffic forecasting task, the RMSE error of the T-GCN and the GRU models are reduced by approximately 50.6$\%$ and 48.8$\%$ compared with the HA model, and the accuracies are approximately 6.8$\%$ and 5.2$\%$ higher than that of HA. The RMSE of the T-GCN and the GRU models are approximately 52.3$\%$ and 50.7$\%$ lower than that of the ARIMA model and the accuracy of these are improved by 41.5$\%$ and 40.4$\%$. Compared with the SVR model, the RMSE of the T-GCN and the GRU models are reduced by 48.0$\%$ and 46.3$\%$, and approximately 4.7$\%$ and 3.0$\%$ higher than that of the SVR model. This is mainly due to methods such as the HA, ARIMA, and SVR that find it difficult to handle complex, nonstationary time series data. The lower prediction effect of the GCN model is because the GCN considers the spatial features only and ignores that the traffic data is typical time series data. In addition, as a mature traffic forecasting method, the ARIMA’s prediction precision is relatively lower than the HA, mainly because the ARIMA has difficulty dealing with long-term and nonstationary data and the ARIMA is calculated by calculating the error of each node and averaging; if there are fluctuations in some data, it will also increase the final total error.

\begin{figure}	
	\centering
	\subfigure[]{
	    \includegraphics[width=0.45\linewidth]{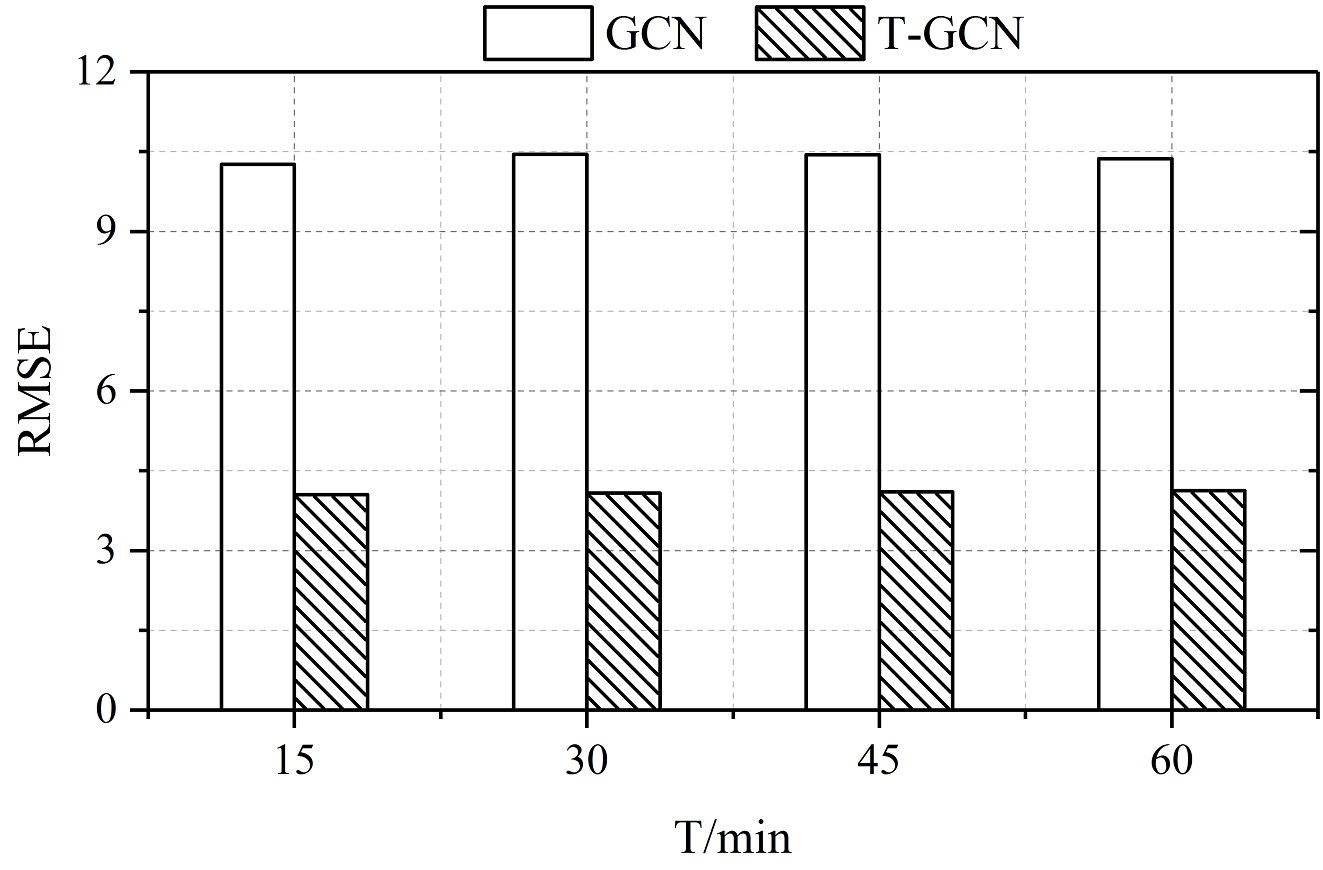}}
	    \label{TGCN-GCN} 
	\subfigure[]{		
		\includegraphics[width=0.45\linewidth]{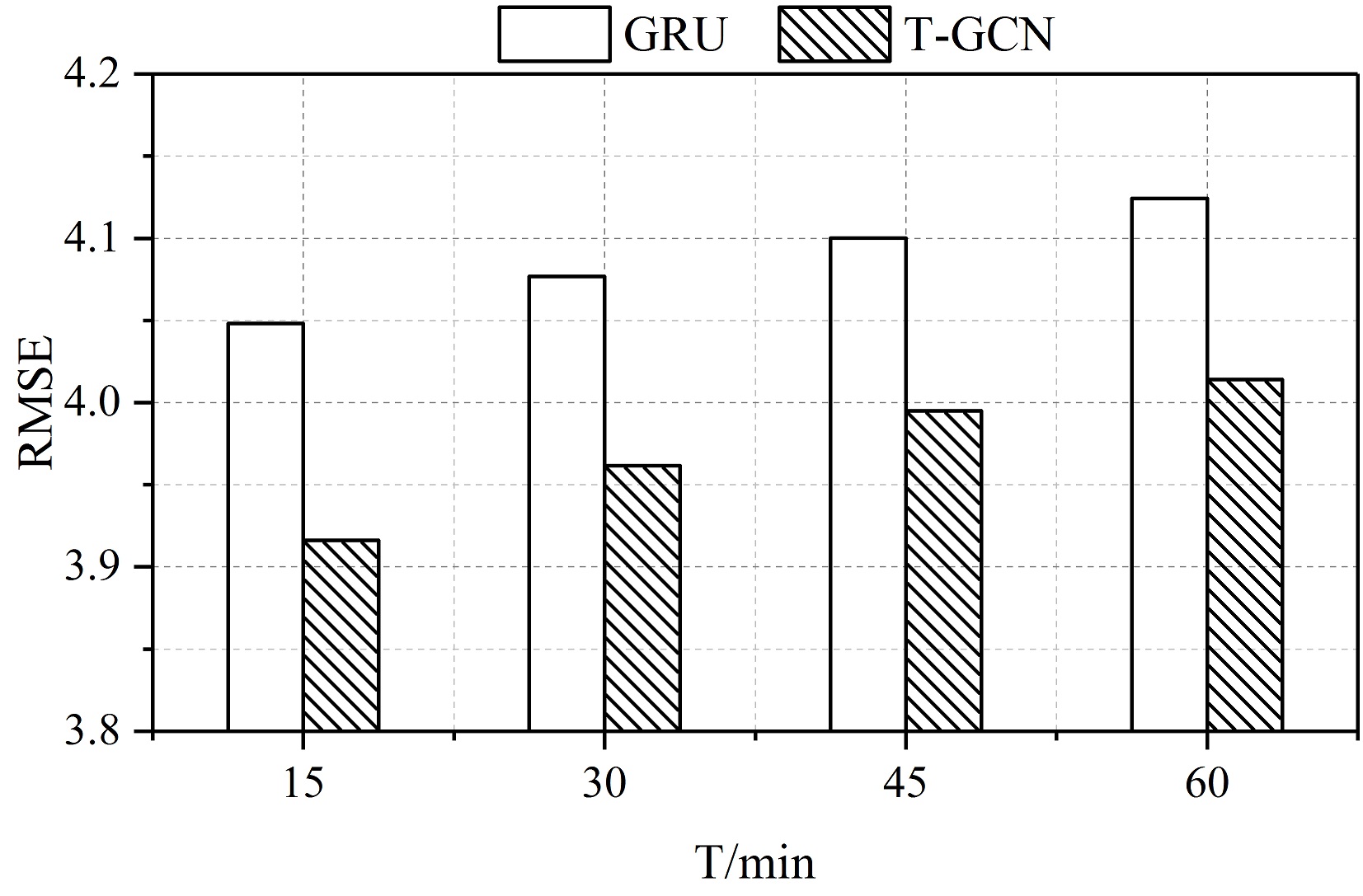}}
		\label{TGCN-GRU} 
	\caption{(a) The RMSE of the T-GCN model lower than the GCN model, which considers spatial feature only, indicating the effectiveness of the T-GCN to capture spatial feature. (b) The RMSE of the T-GCN model lower than the GRU model, which considers temporal feature only, indicating the effectiveness of the T-GCN to capture temporal feature. }
	\label{st-prediction}
\end{figure}

(2) Spatio-temporal prediction capability. To verify whether the T-GCN model has the ability to portray spatial and temporal feature from traffic data, we compare the T-GCN model with the GCN model and the GRU model. As shown in Figure \ref{st-prediction}, we can clearly see that method based on the spatio-temporal features (T-GCN) has better prediction precision than those based on single factor (GCN, GRU), indicating that the T-GCN model can capture spatial and temporal feature from traffic data. For example, for the 15-min traffic forecasting, the RMSE is reduced by approximately 57.8$\%$ compared with the GCN model, which considers only spatial feature and for 30-min traffic forecasting, the RMSE of the T-GCN model is reduced by 57.6$\%$, indicating that the T-GCN model can capture spatial dependence. Compared with the GRU model, which considers only temporal features, for 15-min and 30-min traffic forecasting, the RMSE of the T-GCN model is decreased by approximately 3.3$\%$ and 2.9$\%$, indicating that the T-GCN model can capture temporal dependence well.

\begin{figure}
	\centering
	\subfigure[]{
	    \includegraphics[width=0.9\linewidth]{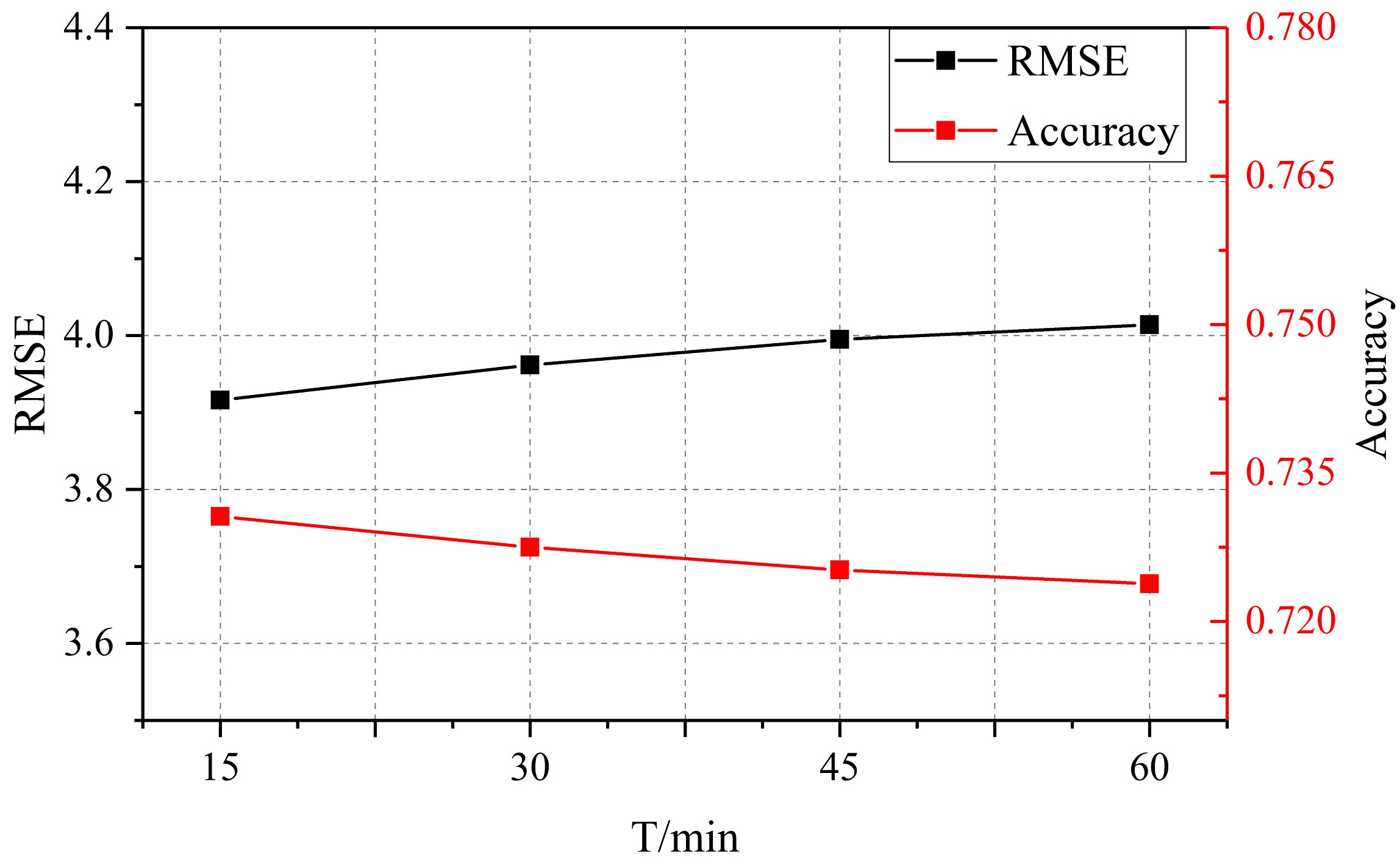}}
	    \label{long-term1} 
	\subfigure[]{		
		\includegraphics[width=0.9\linewidth]{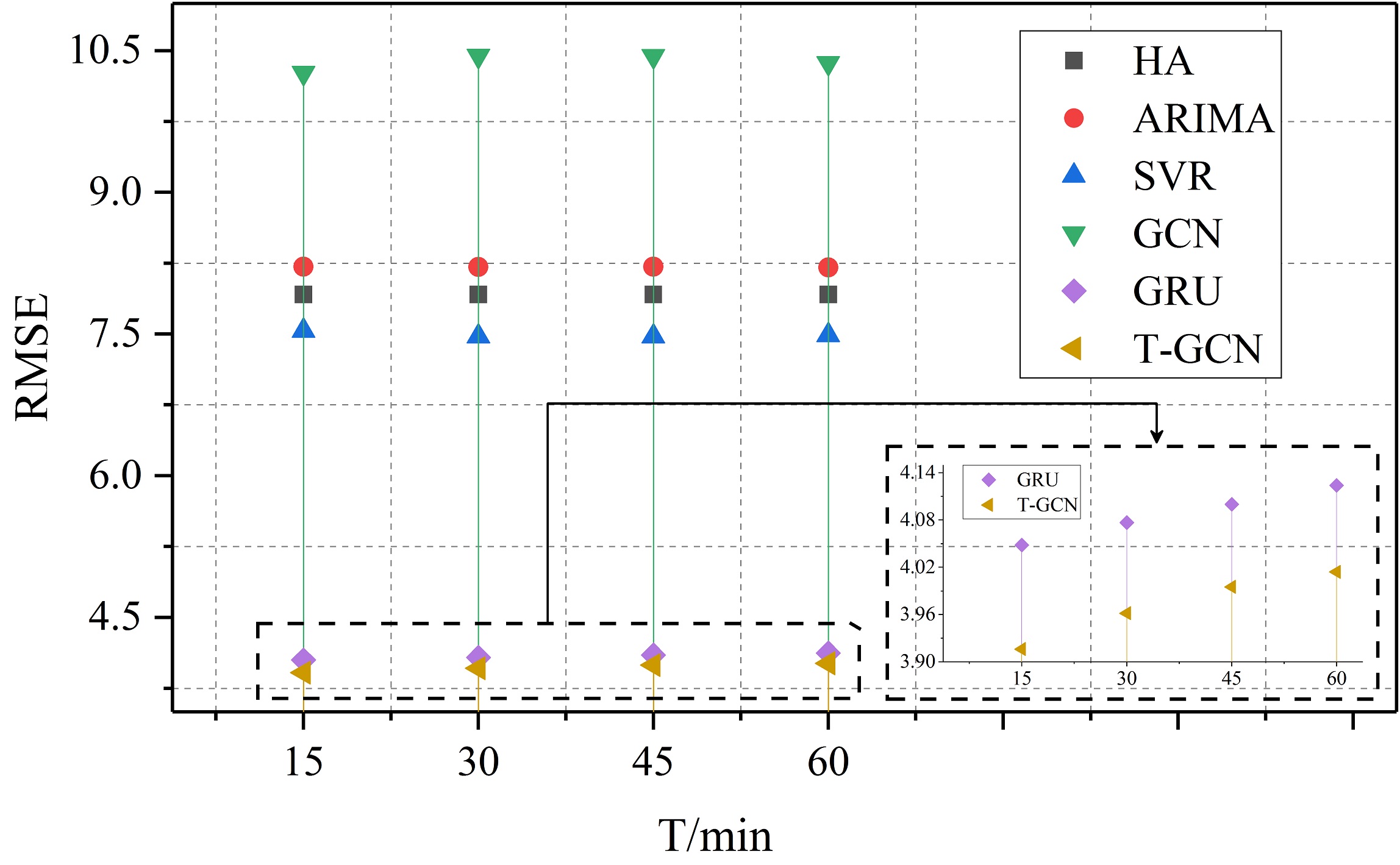}}
		\label{long-term2} 
	\caption{(a) Under different prediction horizons, the change of RMSE and Accuracy are small, indicating that our approach is insensitive to prediction horizons. (b) Under different prediction horizons, the T-GCN model has lowest RMSE error compared to baseline methods. }
	\label{long-term} 
\end{figure}

(3) Long-term prediction ability. No matter how the horizon changes, the T-GCN model can obtain the best prediction performance through training and the prediction results have less tendency to change, indicating that our approach is insensitive to prediction horizons. Thus, we know that the T-GCN model can be used not only for short-term prediction but also for long-term prediction. Figure \ref{long-term}(a) shows the change of RMSE and Accuracy at different prediction horizons, which represent the prediction error and precision of the T-GCN model, respectively. It can be seen that the trends of error increase and precision decrease are small, with a certain degree of stability. Figure \ref{long-term}(b) shows the comparison of RMSE for baselines at different horizons. We observe that the T-GCN model can achieve the best results regardless of the prediction horizon.
\subsection{Perturbation Analysis and Robustness}
There is inevitably noise during the data collection process in the real world. To test the noise immunity of the T-GCN model, we test the robustness of the model through perturbation analysis experiments.

\begin{figure}	
	\centering	
	\subfigure[]{			
		\includegraphics[width=0.45\linewidth]{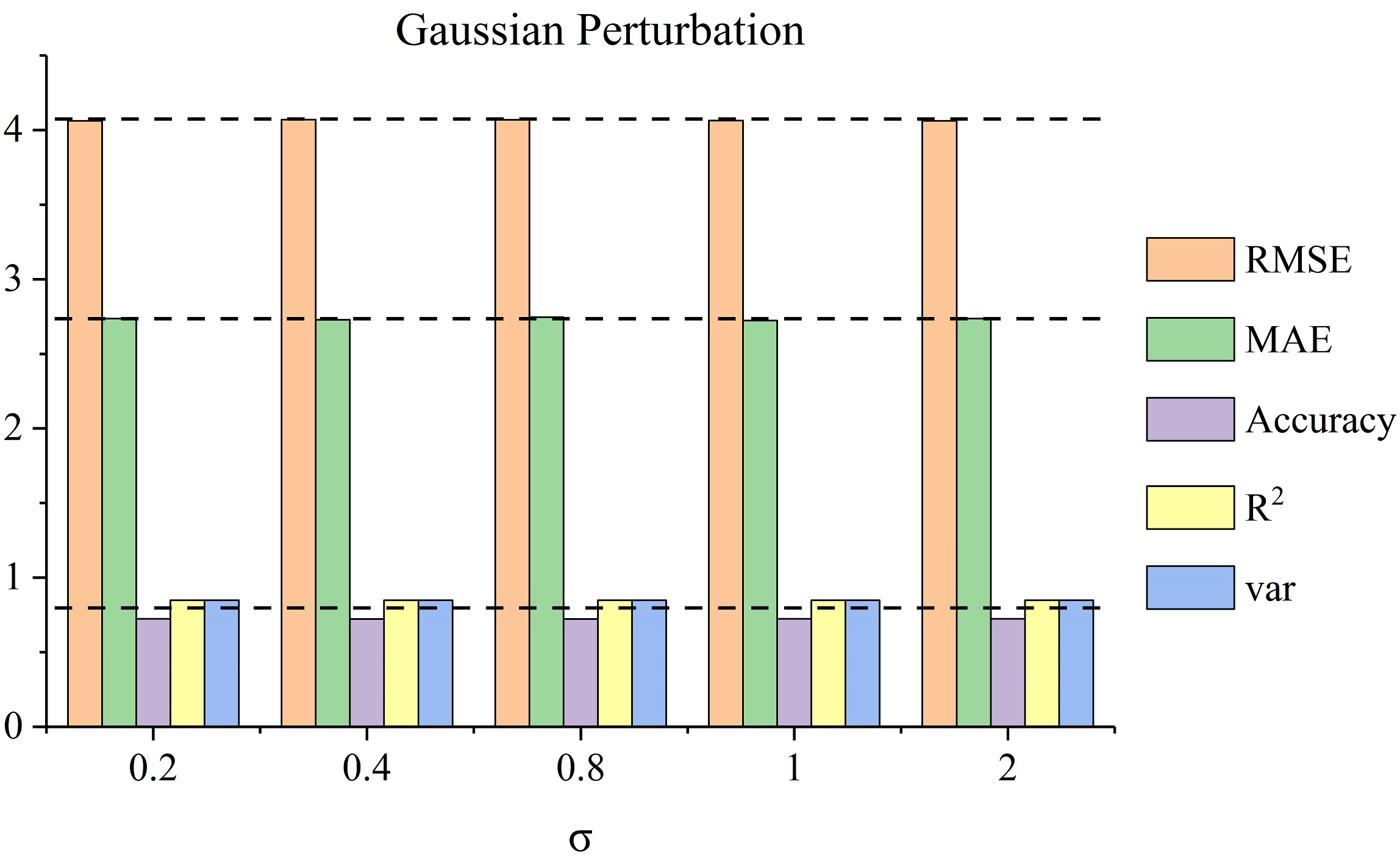}}	
	\label{Gaussian distribution} 
	\subfigure[]{		
		\includegraphics[width=0.45\linewidth]{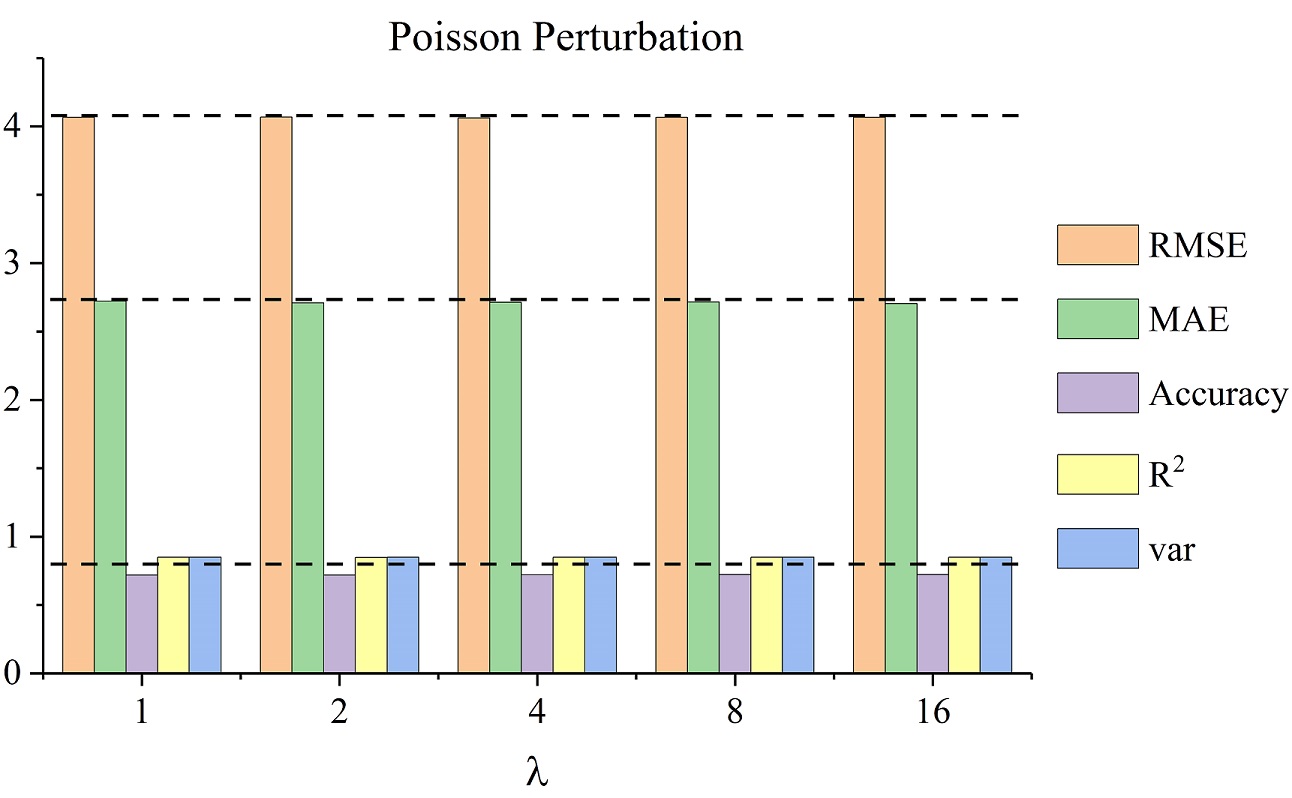}}
	\label{Poisson distribution} 
	\subfigure[]{		
		\includegraphics[width=0.45\linewidth]{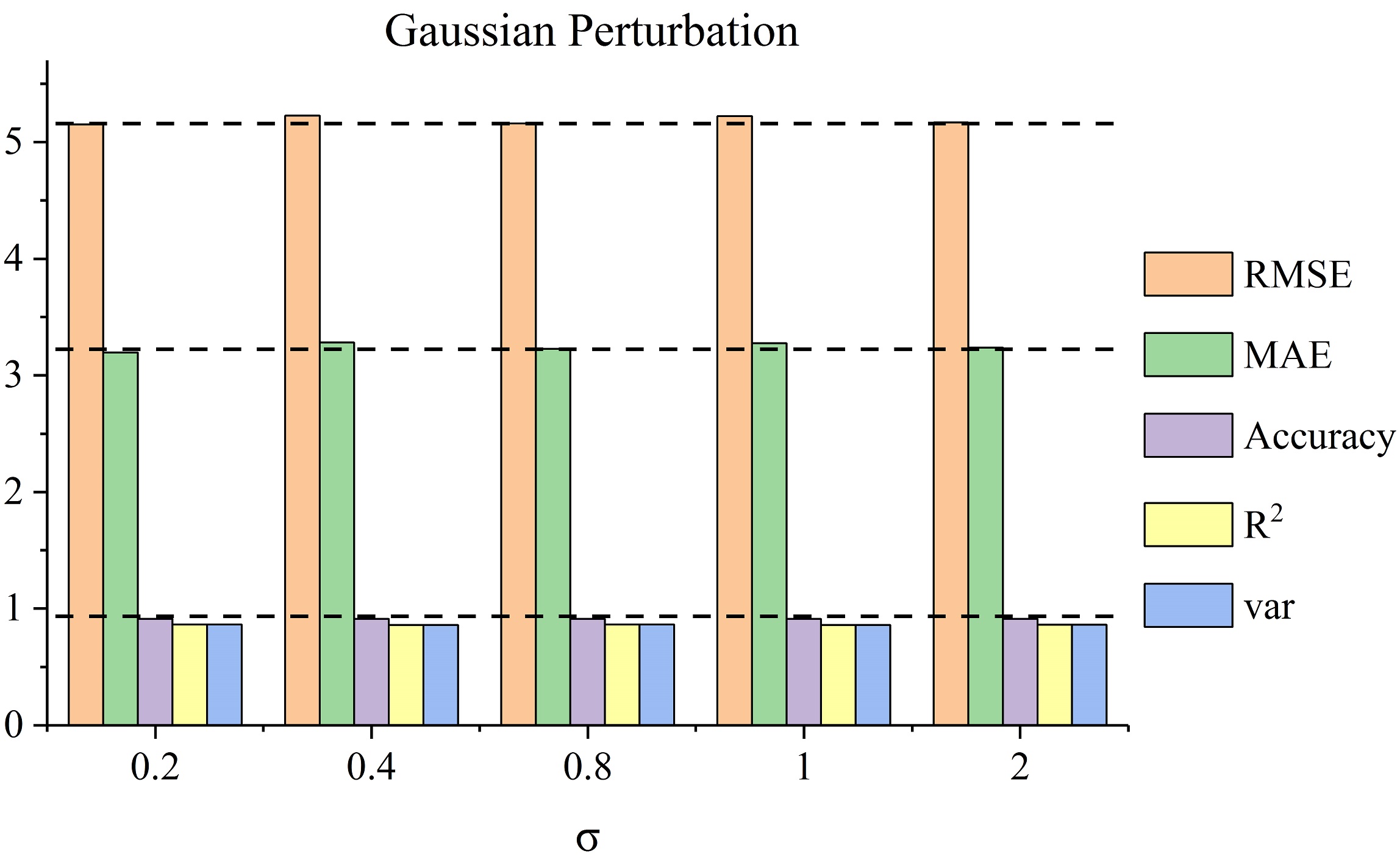}}
	\label{T-GCN-Los-Gaussian} 
	\subfigure[]{		
		\includegraphics[width=0.45\linewidth]{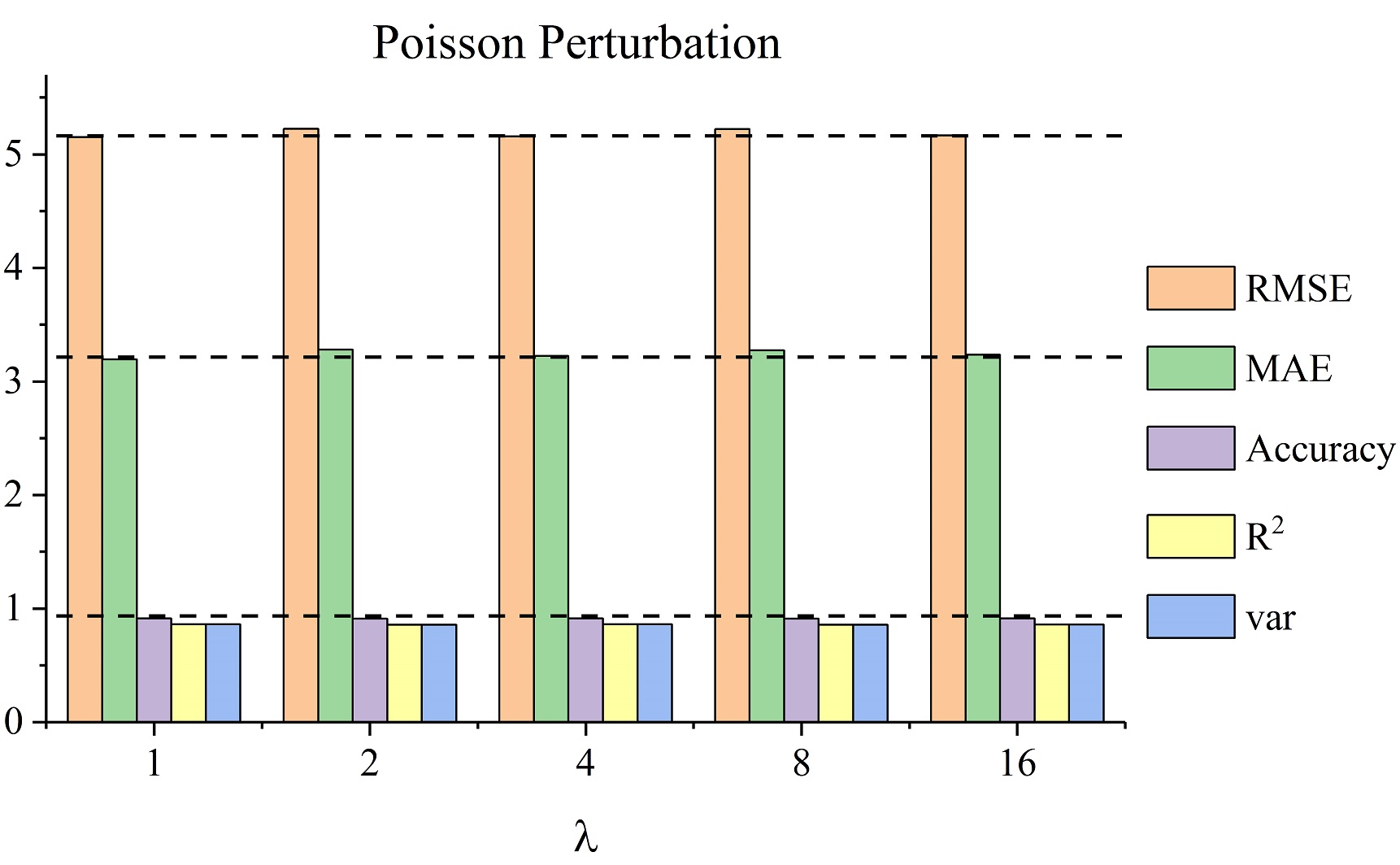}}
	\label{T-GCN-Los-Poisson} 
	\caption{Perturbation analysis. The horizontal axis represents $\sigma$ or $\lambda$, the vertical axis represents prediction results, and different colors mean different metrics. (a) The results of adding Gaussian perturbation on SZ-taxi. (b) The results of adding Poisson perturbation on SZ-taxi. (c) The results of adding Gaussian perturbation on Los-loop. (d) The results of adding Poisson perturbation on Los-loop.}
	\label{Perturbation Analysis} 
\end{figure}

We add two types of commonly random noise to the data during the experiment. The random noise obeys the Gaussian distribution $N\in(0,\sigma^{2})(\sigma \in(0.2,0.4,0.8,1,2))$ and the Poisson distribution $P(\lambda)(\lambda\in(1,2,4,8,16))$ and then we normalize the values of the noise matrices turn to $\left[0,1\right]$. Using different evaluation metrics, the results are shown as following. Figure \ref{Perturbation Analysis}(a) shows the results of adding Gaussian noise on SZ-taxi dataset, where the horizontal axis represents σ, the vertical axis represents the change of each evaluation metrics, and different colors indicate different metrics. Similarly, Figure \ref{Perturbation Analysis}(b) shows the results of adding Poisson noise on SZ-taxi. \ref{Perturbation Analysis}(c) and \ref{Perturbation Analysis}(d) are the results of adding Gaussian noise and Poisson noise based on Los-loop dataset. It can be seen that the metrics change little whatever the noise distribution is. Thus, the T-GCN model is robust and is able to handle high noise issues.
\subsection{Model Interpretation}
\begin{figure}
    \centering
    \subfigure{\includegraphics[width=0.9\linewidth]{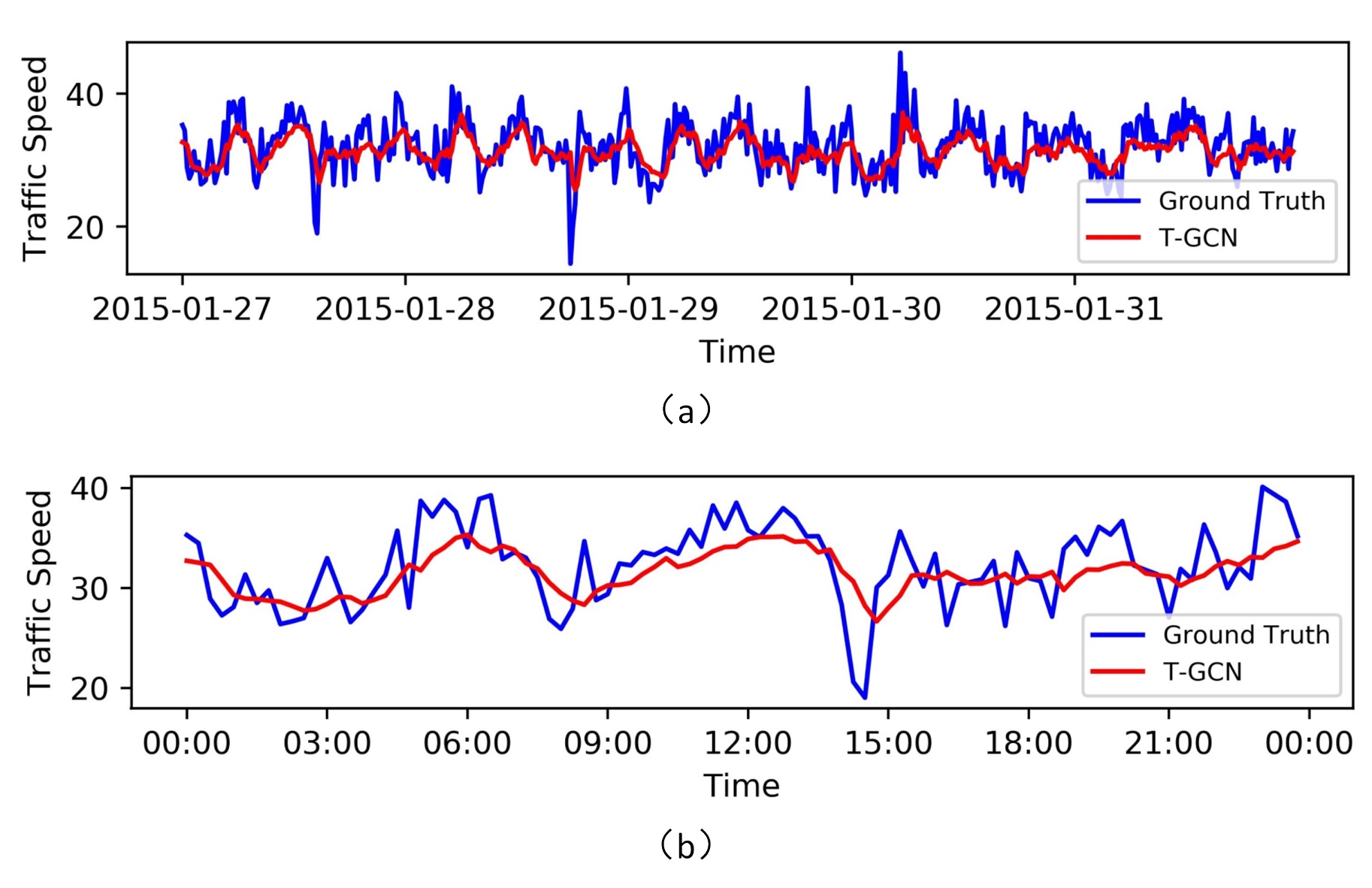}}
	\caption{The visualization results for prediction horizon of 15 minutes.}
	\label{15min}
    \subfigure{\includegraphics[width=0.9\linewidth]{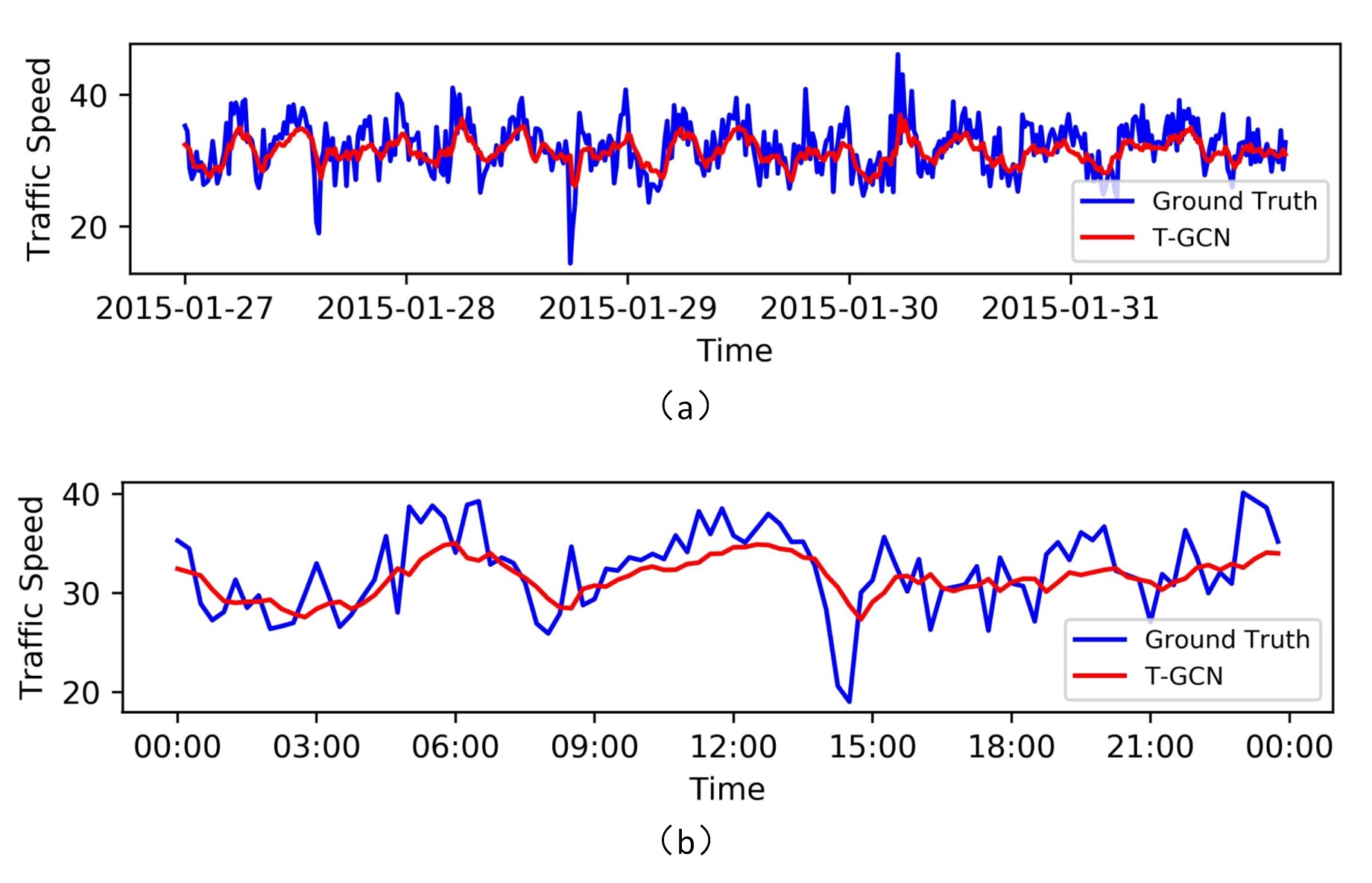}}
	\caption{The visualization results for prediction horizon of 30 minutes.}
	\label{30min}
	\subfigure{\includegraphics[width=0.9\linewidth]{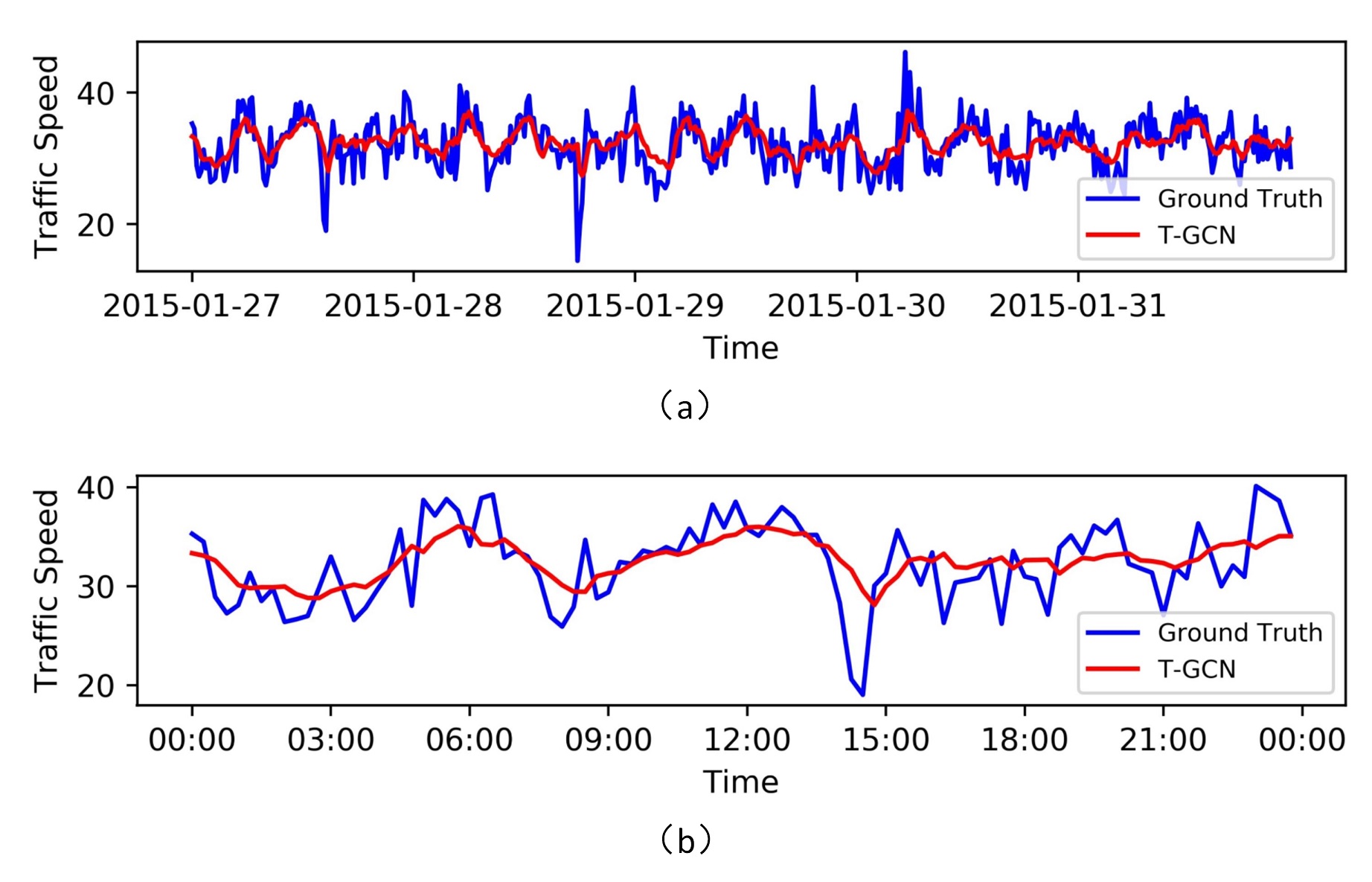}}
	\caption{The visualization results for prediction horizon of 45 minutes.}
	\label{45min}
	\subfigure{\includegraphics[width=0.9\linewidth]{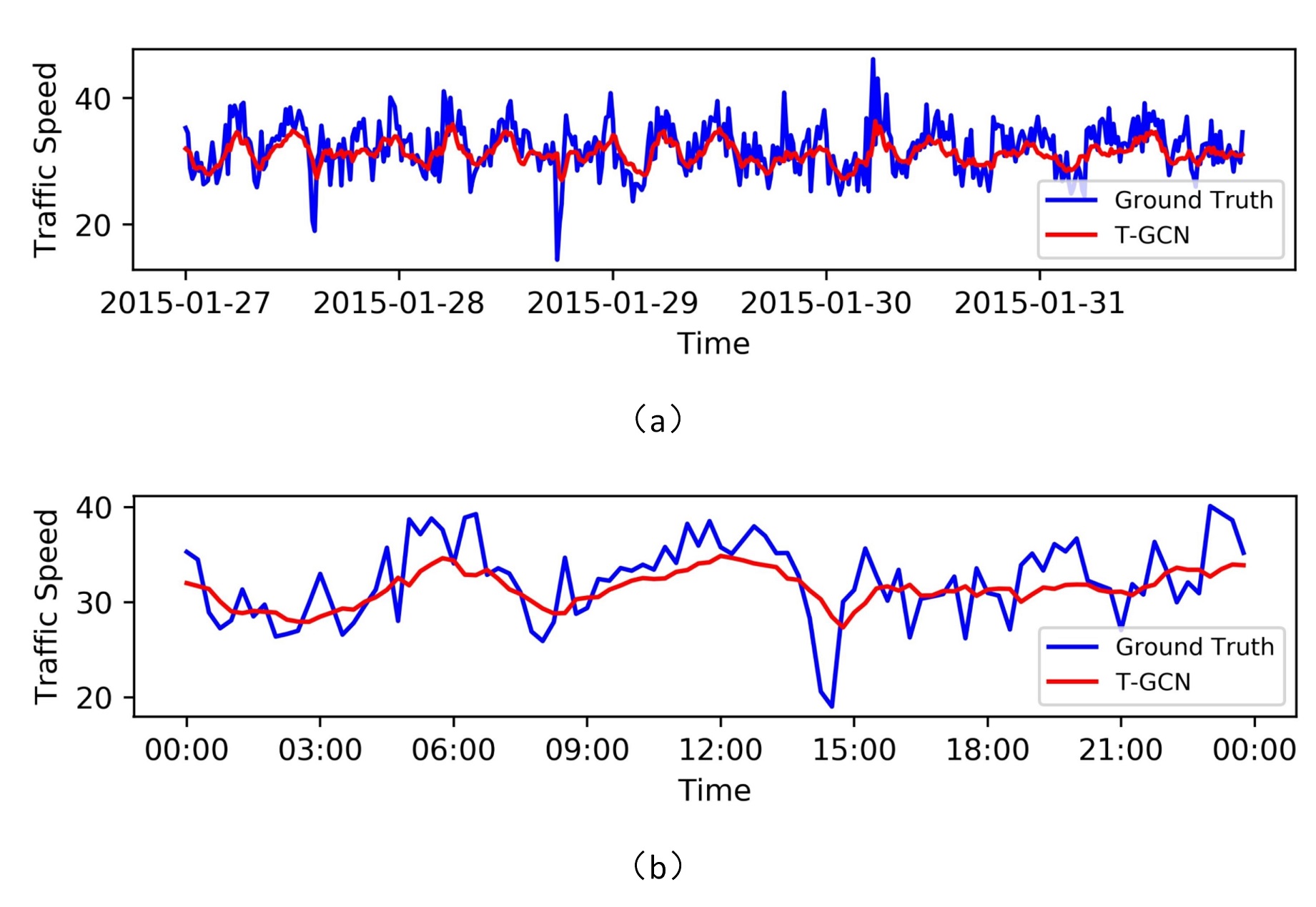}}
	\caption{The visualization results for prediction horizon of 60 minutes.}
	\label{60min}
\end{figure}

To better understand the T-GCN model, we select one road on SZ-taxi dataset and visualize prediction results of the test set. Figure \ref{15min}, Figure \ref{30min}, Figure \ref{45min}, and Figure \ref{60min} show the visualization results for prediction horizons of 15 minutes, 30 minutes, 45 minutes, and 60 minutes, respectively. These results show: 

(1) The T-GCN model predicts poorly at the peak. We speculate that the main cause is that the GCN model defines a smooth filter in the Fourier domain and captures spatial feature by constantly moving the filter. This process leads to a small change in the overall prediction results, which makes the peak smoother. 

(2) There is a certain error between the real traffic information and the prediction results. One error is mainly because when there are no taxis on the roads, there will be no information record with a zero value. The other error is because when the traffic information value is small, a small difference can cause a big relative error.

(3) Regardless of the prediction horizons, the T-GCN model can always achieve better results. The T-GCN model can capture the spatio-temporal features and obtain the variation trend of traffic information on the road. Moreover, The T-GCN model detect the start and end of the rush hour and make prediction results with similar pattern with the real traffic speed. Those properties are helpful for predicting traffic congestion and other traffic phenomena.

\section{Conclusion}
In this paper, we propose a novel neural network-based approach for traffic forecasting called T-GCN, which combines the GCN and the GRU. We use a graph network to model the urban road network in which the nodes on the graph represent roads, the edges represent the connection relationships between roads, and the traffic information on the roads is described as the attribute of the nodes on the graph. On one hand, the GCN is used to capture the topological structure of the graph to obtain the spatial dependence; on the other hand, the GRU model is used to capture the dynamic change of node attribute to obtain the temporal dependence. Eventually the T-GCN model is used to tackle spatio-temporal traffic forecasting tasks. When evaluated on two real-world traffic datasets and compared with the HA model, the ARIMA model, the SVR model, the GCN model, and the GRU model, the T-GCN model achieves the best prediction results under different prediction horizons. In addition, the perturbation analysis illustrates the robustness of our approach. In summary, the T-GCN model can successfully capture the spatial and temporal features from traffic data and is not limited to traffic forecasting, but can also be applied to other spatio-temporal tasks.


%



\ifCLASSOPTIONcompsoc
  \section*{Acknowledgments}
\else
  \section*{Acknowledgment}
\fi

This work was supported by the National Science Foundation of China [grant numbers 41571397, 41501442, and 51678077], and by Natural Science Foundation of Hunan Province (2016JJ3144 and 2016JJ2006).

\ifCLASSOPTIONcaptionsoff
  \newpage
\fi


\bibliographystyle{IEEEtran}
\bibliography{MyReference}

\begin{thebibliography}{1}

\bibitem{Huang2005Dyanamic}
H.~Huang, ``Dynamic modeling of urban transportation networks and analysis of its travel behaviors,'' \emph{Chinese Journal of Management}, vol.~2, pp. 18--22, Jan. 2005.

\bibitem{Liu2004A}
J.~Liu and W.~Guan, ``A summary of traffic flow forecasting methods,'' \emph{Journal of Highway Transportation Research Development}, Mar. 2004.

\bibitem{Yuan2012Synthesis}
J.~Yuan and B.~Fan, ``Synthesis of short-term traffic flow forecasting research progress,'' \emph{Urban Transport of China}, Jun. 2012.  

\bibitem{Dong2012Spatial}
C.~J. Dong, C.~F. Shao, Z.~Cheng-Xiang, and M.~Meng, ``Spatial and temporal characteristics for congested traffic on urban expressway,'' \emph{Journal of	Beijing University of Technology}, vol.~38, no.~8, pp. 1242--1246+1268, 2012.

\bibitem{Ahmed1979ANALYSIS}
M.~S. Ahmed and A.~R. Cook, ``Analysis of freeway traffic time-series data by using box-jenkins techniques,'' \emph{Transportation Research Board}. no.~722, pp. 1-9, 1979.

\bibitem{Hamed1995Short}
M.~M. Hamed, H.~R. Al-Masaeid, and Z.~M.~B. Said, ``Short-term prediction of traffic volume in urban arterials,'' \emph{Journal of Transportation Engineering}, vol. 121, no.~3, pp. 249--254, 1995.

\bibitem{Okutani1984Dynamic}
I.~Okutani and Y.~J. Stephanedes, ``Dynamic prediction of traffic volume through kalman filtering theory,'' \emph{Transportation Research Part B Methodological}, vol.~18, no.~1, pp. 1--11, 1984.

\bibitem{Wu2004Travel}
C.~H. Wu, J.~M. Ho, and D.~T. Lee, ``Travel-time prediction with support vector regression,'' \emph{IEEE Transactions on Intelligent Transportation Systems}, vol.~5, no.~4, pp. 276--281, Dec. 2004.

\bibitem{Yao2006Research}
Z.~S. Yao, C.~F. Shao, and Y.~L. Gao, ``Research on methods of short-term traffic forecasting based on support vector regression,'' \emph{Journal of Beijing Jiaotong University}, vol.~30, no.~3, pp. 19--22, 2006.

\bibitem{Zhang2009Short}
X.~L. Zhang, H.~E. Guo-Guang, and L.~U. Hua-Pu, ``Short-term traffic flow forecasting based on k-nearest neighbors non-parametric regression,'' \emph{Journal of Systems Engineering}, vol.~24, no.~2, pp. 178--183, Feb. 2009

\bibitem{Sun2006A}
S.~Sun, C.~Zhang, and G.~Yu, ``A bayesian network approach to traffic flow forecasting,'' \emph{IEEE Transactions on Intelligent Transportation Systems}, vol.~7, no.~1, pp. 124--132, Mar. 2006.

\bibitem{Huang2014Deep}
W.~Huang, G.~Song, H.~Hong, and K.~Xie, ``Deep architecture for traffic flow prediction: Deep belief networks with multitask learning,'' \emph{IEEE Transactions on Intelligent Transportation Systems}, vol.~15, no.~5, pp. 2191--2201, Apr. 2014.

\bibitem{Fu2017Using}
R.~Fu, Z.~Zhang, and L.~Li, ``Using lstm and gru neural network methods for traffic flow prediction,'' \emph{Youth Academic Conference of Chinese Association of Automation}, pp. 324--328, Jan. 2017.

\bibitem{Zhang2016Deep}
J.~Zhang, Y.~Zheng, and D.~Qi, ``Deep spatio-temporal residual networks for citywide crowd flows prediction,'' Sep. 2016.

\bibitem{Wu2016Short}
Y.~Wu and H.~Tan, ``Short-term traffic flow forecasting with spatial-temporal correlation in a hybrid deep learning framework,'' Dec. 2016.

\bibitem{Cao2017Interactive}
X.~Cao, Y.~Zhong, Y.~Zhou, J.~Wang, C.~Zhu, and W.~Zhang, ``Interactive temporal recurrent convolution network for traffic prediction in data centers,'' \emph{IEEE Access}, vol.~PP, no.~99, pp. 1--1, 2017.

\bibitem{Defferrard2016Convolutional}
M.~Defferrard, X.~Bresson, and P.~Vandergheynst, ``Convolutional neural networks on graphs with fast localized spectral filtering,'' Jun. 2016.

\bibitem{Xu2014Analysis}
X.~Y. Xu, J.~Liu, H.~Y. Li, and J.~Q. Hu, ``Analysis of subway station capacity with the use of queueing theory,'' \emph{Transportation Research Part C Emerging Technologies}, vol.~38, no.~1, pp. 28--43,  Jan. 2014.

\bibitem{Wei2013Total}
P.~Wei, Y.~Cao, and D.~Sun, ``Total unimodularity and decomposition method for large-scale air traffic cell transmission model,'' \emph{Transportation Research Part B}, vol.~53, no.~3, pp. 1--16, Jul. 2013.

\bibitem{Qi2011Traffic}
W.~Qi, L.~I. Li, H.~U. Jianming, and B.~Zou, ``Traffic velocity distributions for different spacings,'' \emph{Journal of Tsinghua University}, vol.~51, no.~3, pp. 309--312, Mar. 2011.

\bibitem{Xu2013Impacts}
F.~F. Xu, Z.~C. He, and Z.~R. Sha, ``Impacts of traffic management measures on urban network microscopic fundamental diagram,'' \emph{Journal of Transportation Systems Engineering and Information Technology}, vol.~13, no.~2, pp. 185--190,  Apr. 2013.

\bibitem{Vlahogianni2015Computational}
E.~I. Vlahogianni, ``Computational Intelligence and Optimization for Transportation Big Data: Challenges and Opportunitie,'' \emph{Springer International Publishing}, pp. 107-128, May. 2015.

\bibitem{Shan2013Urban}
Z.~Shan, D.~Zhao, and Y.~Xia, ``Urban road traffic speed estimation for missing probe vehicle data based on multiple linear regression model,'' \emph{16th International IEEE Conference on Intelligent Transportation Systems}, pp. 118--123, 2013.

\bibitem{Shen2011Short}
G.~J. Shen, X.~H. Wang, and X.~J. Kong, ``Short-term traffic volume intelligent hybrid forecasting model and its application,'' \emph{Systems Engineering-Theory and Practice}, vol.~31, no.~3, pp. 561--568, 2011.

\bibitem{Eleni2004Short}
E.~I. Vlahogianni, J.~C. Golias, and M.~G. Karlaftis, ``Short‐term traffic forecasting: Overview of objectives and methods,'' \emph{Transport Reviews}, vol.~24, no.~5, pp. 533--557, Sep. 2004.

\bibitem{Van2012Short}
H.~Van~Lint and C.~Van~Hinsbergen, ``Short-term traffic and travel time prediction models,'' \emph{Transportation Research E-Circular}, pp. 22--41, Nov. 2012.


\bibitem{Sun2004Interval}
H.~Sun, C.~Zhang, and B.~Ran, ``Interval prediction for traffic time series using local linear predictor,'' \emph{The International IEEE Conference on Intelligent Transportation Systems}, pp. 410--415, Nov. 2004.

\bibitem{Dudek2016Pattern}
G.~Dudek, ``Pattern-based local linear regression models for short-term load forecasting,'' \emph{Electric Power Systems Research}, vol. 130, pp. 139--147, Jan. 2016.

\bibitem{Voort1996Combining}
M.~V.~D. Voort, M.~Dougherty, and S.~Watson, ``Combining kohonen maps with arima time series models to forecast traffic flow,'' \emph{Transportation Research Part C Emerging Technologies}, vol.~4, no.~5, pp. 307--318, Oct. 1996.

\bibitem{Lee1999Application}
S.~Lee and D.~Fambro, ``Application of subset autoregressive integrated moving average model for short-term freeway traffic volume forecasting,'' \emph{Transportation Research Record Journal of the Transportation Research Board}, vol.~1678, no.~1, pp. 179--188, 1999.

\bibitem{Fabian2003Modeling}
X.~Fabian, G.~Ban, R.~Boussaïd, M.~Breitenfeldt, C.~Couratin, P.~Delahaye, D.~Durand, P.~Finlay, X.~Fléchard, and B.~Guillon, ``Modeling and forecasting vehicular traffic flow as a seasonal arima process: Theoretical basis and empirical results,'' \emph{Journal of Transportation Engineering}, vol. 129, no.~6, pp. 664--672, Feb. 2003.

\bibitem{Lippi2013Short}
M.~Lippi, M.~Bertini, and P.~Frasconi, ``Short-term traffic flow forecasting: An experimental comparison of time-series analysis and supervised learning,'' \emph{IEEE Transactions on Intelligent Transportation Systems}, vol.~14, no.~2, pp. 871--882, Mar. 2013.

\bibitem{Hinsbergen2012Localized}
C.~P. I. J.~V. Hinsbergen, T.~Schreiter, F.~S. Zuurbier, J.~W. C.~V. Lint, and H.~J.~V. Zuylen, ``Localized extended kalman filter for scalable real-time traffic state estimation,'' \emph{IEEE Transactions on Intelligent Transportation Systems}, vol.~13, no.~1, pp. 385--394,  Mar. 2012.

\bibitem{Ojeda2013Adaptive}
L.~L. Ojeda, A.~Y. Kibangou, and C.~C. de~Wi, ``Adaptive kalman filtering for multi-step ahead traffic flow prediction,'' \emph{2013 American Control Conference},  Aug. 2013.

\bibitem{Smola2004A}
A.~J. Smola and B.~Schölkopf, ``A tutorial on support vector regression,'' \emph{Statistics and Computing}, vol.~14, no.~3, pp. 199--222, Jan. 2004.

\bibitem{Yin2002Urban}
H.~Yin, S.~C. Wong, J.~Xu, and C.~K. Wong, ``Urban traffic flow prediction using a fuzzy-neural approach,'' \emph{Transportation Research Part C}, vol.~10, no.~2, pp. 85--98, Apr. 2002.

\bibitem{Silver2016Mastering}
D.~Silver, A.~Huang, C.~J. Maddison, A.~Guez, L.~Sifre, G.~V.~D. Driessche, J.~Schrittwieser, I.~Antonoglou, V.~Panneershelvam, and M.~Lanctot, ``Mastering the game of go with deep neural networks and tree search,'' \emph{Nature}, vol. 529, no. 7587, pp. 484--489, Jan. 2016.

\bibitem{Silver2017Mastering}
D.~Silver, J.~Schrittwieser, K.~Simonyan, I.~Antonoglou, A.~Huang, A.~Guez, T.~Hubert, L.~Baker, M.~Lai, and A.~Bolton, ``Mastering the game of go without human knowledge,'' \emph{Nature}, vol. 550, no. 7676, pp. 354--359, Oct. 2017.

\bibitem{Morav2017DeepStack}
M.~Moravčík, M.~Schmid, N.~Burch, V.~Lisý, D.~Morrill, N.~Bard, T.~Davis, K.~Waugh, M.~Johanson, and M.~Bowling, ``Deepstack: Expert-level artificial intelligence in heads-up no-limit poker,'' \emph{Science}, vol. 356, no. 6337, p. 508, Jan. 2017.

\bibitem{Park2010Forecasting}
D.~Park and L.~R. Rilett, ``Forecasting freeway link travel times with a multilayer feedforward neural network,'' \emph{Computer-Aided Civil and Infrastructure Engineering}, vol.~14, no.~5, pp. 357--367, Dec. 2010.

\bibitem{J2002FREEWAY}
H.~J. v.~Z. J.~W. C.~van Lint, S. P.~Hoogendoorn, S.~P. Hoogendoorn, and H.~J.~V. Zuylen, ``FREEWAY TRAVEL TIME PREDICTION WITH STATE-SPACE NEURAL NETWORKS: MODELING STATE-SPACE DYNAMICS WITH RECURRENT NEURAL NETWORKS,'' \emph{Transportation Research Record}, vol.~1811, no.~1, pp. 347--369, Jan. 2002.

\bibitem{Lv2015Traffic}
Y.~Lv, Y.~Duan, W.~Kang, Z.~Li, and F.~Y. Wang, ``Traffic flow prediction with big data: A deep learning approach,'' \emph{IEEE Transactions on Intelligent Transportation Systems}, vol.~16, no.~2, pp. 865--873, Jan. 2015.

\bibitem{Ke2017Short}
J.~Ke, H.~Zheng, H.~Yang, Xiqun, and Chen, ``Short-term forecasting of passenger demand under on-demand ride services: A spatio-temporal deep learning approach,'' \emph{Transportation Research Part C: Emerging Technologies}. vol.~85, pp. 591--608,Jun. 2017.

\bibitem{Yu2017spatio-temporal}
H.~Yu, Z.~Wu, S.~Wang, Y.~Wang, and X.~Ma, ``spatio-temporal recurrent convolutional networks for traffic prediction in transportation networks,'' \emph{Sensors}, vol.~17, no.~7, Jun. 2017.

\bibitem{Kipf2016Semi}
T.~N. Kipf and M.~Welling, ``Semi-supervised classification with graph convolutional networks,'' Sep. 2016.

\bibitem{Li2017Graph}
Y.~Li, R.~Yu, C.~Shahabi, and Y.~Liu, ``Graph convolutional recurrent neural network: Data-driven traffic forecasting,'' Jul. 2017.

\bibitem{Bruna2014Spectral}
J.~Bruna, W.~Zaremba, A.~Szlam, and Y.~Lecun, ``Spectral networks and locally connected networks on graphs,'' \emph{Computer Science}, Dec. 2013.

\bibitem{Bengio2002Learning}
Y.~Bengio, P.~Simard, and P.~Frasconi, ``Learning long-term dependencies with gradient descent is difficult,'' \emph{IEEE Trans Neural Networks}, vol.~5, no.~2, pp. 157--166, 2002.

\bibitem{Sepp1997Long}
J.~S. Sepp~Hochreiter, ``Long short-term memory,'' \emph{Neural Computation}, vol.~9, no.~8, pp. 1735--1780, Dec. 1997.

\bibitem{Cho2014On}
K.~Cho, B.~V. Merrienboer, D.~Bahdanau, and Y.~Bengio, ``On the properties of neural machine translation: Encoder-decoder approaches,'' \emph{Computer Science}, Sep. 2014.

\bibitem{Chung2014Empirical}
J.~Chung, C.~Gulcehre, K.~H. Cho, and Y.~Bengio, ``Empirical evaluation of gated recurrent neural networks on sequence modeling,'' \emph{Eprint Arxiv}, Dec. 2014.

\end{thebibliography}

%


%






\end{document}